\documentclass[runningheads]{llncs}
\usepackage[T1]{fontenc}
\usepackage{graphicx}
\usepackage{amssymb}
\usepackage[english]{babel}
\usepackage[all]{foreign}
\usepackage{multirow}
\usepackage{paralist}
\usepackage[inline]{enumitem} 
\usepackage{subfig}
\usepackage{tikz}

\tikzset{every picture/.style={line width=0.75pt}}
\usetikzlibrary{arrows,positioning,calc}

\usepackage{xifthen}
\usepackage{mathtools}
\usepackage{tikz-cd}
\usetikzlibrary{decorations.pathmorphing}

\newcommand{\atom}[1]{#1}
\newcommand{\poss}[1]{\mathsf{#1}}
\newcommand{\pem}[1]{\mathsf{#1}}

\newcommand{\atomSet}{\mathcal{P}}
\newcommand{\agentSet}{\mathcal{AG}}
\newcommand{\actionSet}{\mathcal{A}}

\newcommand{\E}{\mathcal{E}}
\newcommand{\Lang}[1]{\mathcal{L}_{\atomSet, \agentSet}^{#1}}

\newcommand{\mal}{m\mathcal{A}^*}

\newcommand{\asp}[1]{\mathtt{#1}}

\newcommand{\planex}[2]{\textnormal{\textsc{PlanEx}}(#1, #2)}
\newcommand{\textover}[2]{\mathrel{\overset{\makebox[0pt]{\mbox{\normalfont\tiny\sffamily #2}}}{\ #1\ }}}
\newcommand{\mypar}[1]{\noindent\textbf{#1}}

\newcommand\bisim{\underline{\leftrightarrow}}

\newcommand\utimes{\mathbin{\ooalign{$\cup$\cr%
   \hfil\raise0.42ex\hbox{$\scriptscriptstyle\times$}\hfil\cr}}}
\newcommand\bigutimes{\mathop{\ooalign{$\bigcup$\cr%
   \hfil\raise0.36ex\hbox{$\scriptscriptstyle\boldsymbol{\times}$}\hfil\cr}}}

\begin{document}
    \title{\textsc{delphic}: Practical DEL Planning via Possibilities (Extended Version)}
    
    
    \author{
        Alessandro Burigana\inst{1}\orcidID{0000-0002-9977-6735} \and
        Paolo Felli\inst{2}\orcidID{0000-0001-9561-8775} \and
        Marco Montali\inst{1}\orcidID{0000-0002-8021-3430}
    }
    
    \authorrunning{A. Burigana et al.}
    
    \institute{
        Free University of Bozen-Bolzano, Italy
        \email{\{burigana, montali\}@inf.unibz.it}
        \and
        University of Bologna, Italy
        \email{paolo.felli@unibo.it}
    }
    
    \maketitle              
    
    \begin{abstract}
    Dynamic Epistemic Logic (DEL) provides a framework for epistemic planning that is capable of representing non-deterministic actions, partial observability, higher-order knowledge and both factual and epistemic change. The high expressivity of DEL challenges existing epistemic planners, which typically can handle only restricted fragments of the whole framework. 
    The goal of this work is to push the envelop of practical DEL planning, ultimately aiming for epistemic planners to be able to deal with the full range of features offered by DEL. Towards this goal, we question the traditional semantics of DEL, defined in terms on Kripke models. In particular, we propose an equivalent semantics defined using, as main building block, so-called \emph{possibilities}: non well-founded objects representing both factual properties of the world, and what agents consider to be possible. We call the resulting framework \textsc{delphic}. We argue that \textsc{delphic} indeed provides a more compact representation of epistemic states. To substantiate this claim, we implement both approaches in ASP and we set up an experimental evaluation to compare \textsc{delphic} with the traditional, Kripke-based approach. The evaluation confirms that \textsc{delphic} outperforms the traditional approach in space and time.
\end{abstract}

    \section{Introduction}
    Multiagent Systems are employed in a wide range of settings, where autonomous agents are expected to face dynamic situations and to be able to adapt in order to reach a given goal. In these contexts, it is crucial for agents to be able to reason on their physical environment as well as on the \emph{knowledge} that they have about other agents and the knowledge those possess. 
    
    Bolander and Andersen \cite{journals/jancl/Bolander2011} introduced \emph{epistemic planning} as a planning framework based on Dynamic Epistemic Logic (DEL), where \emph{epistemic states} are represented as Kripke models, \emph{event models} are used for representing epistemic actions, and \emph{product updates} define the application of said actions on states. On the one hand, the resulting framework is very expressive, and it allows one to naturally represent non-deterministic actions, partial observability of agents, higher-order knowledge and both factual and epistemic changes. 
    On the other hand, decidability of epistemic planning is not guaranteed in general~\cite{journals/jancl/Bolander2011}. This has led to a considerable body of research adopting the DEL framework to obtain (un)decidability results for fragments of the epistemic planning problem (see \cite{journals/ai/Bolander2020} for a detailed exposition), typically by constraining the event models of actions. Nonetheless, even when such restriction are in place, epistemic planners directly employing the Kripke-based semantics of possible worlds face high complexities, hence considerable efforts have been put in studying action languages that are more amenable computationally \cite{journals/corr/Baral2015,conf/aips/Kominis2015,conf/icaps/Fabiano2020}.
    
    In contrast with the traditional approach in the literature, in this paper we depart from the Kripke-based semantics for DEL and adopt an alternative representation called \emph{possibilities}, first introduced by Gerbrandy and Groeneveld \cite{journals/jolli/Gerbrandy1997}. As we are going to show experimentally, this choice is motivated primarily by practical considerations. In fact, as we expand in Section \ref{sec:delphic}, possibilities support a concise representation of factual and epistemic information and yield a light update operator that promises to achieve better performances compared to the traditional Kripke-based semantics. This is due to the fact that possibilities are \emph{non-well-founded objects}, namely objects that have a \emph{circular} representation (see Aczel \cite{books/csli/Aczel1988} for an exhaustive introduction on non-well-founded set theory). In fact, due to their non-well-founded nature, possibilities naturally reuse previously calculated information, thus drastically reducing the computational overhead deriving from redundant information. Conceptually, whenever an agent does not update his knowledge upon an action, then the possibilities representing its knowledge are directly reused (see Examples \ref{ex:product-update} and \ref{ex:union-update}).
    
    This paper presents a novel formalization of epistemic planning based on possibilities. Although these objects have been previously used in place of Kripke models to represent epistemic states \cite{conf/icaps/Fabiano2020}, previous semantics lacked a general characterization of actions. In this paper, we complement the possibility-based representation of states by formalizing two novel concepts: \emph{eventualities}, representing epistemic actions, and \emph{union update}, providing an update operator based on possibilities and eventualities.
    The resulting planning framework, called \textsc{delphic} (\emph{DEL-planning with a Possibility-based Homogeneous Information Characterisation}), benefits from the compactness of possibilities and promises to positively impact the performance of planning.
    This suggests that \textsc{delphic} is a viable but also convenient alternative to Kripke-based representations. 
    We support this claim by implementing both frameworks in ASP and by setting up an experimental evaluation of the two implementations aimed at comparing the traditional Kripke semantics for DEL and \textsc{delphic}. The comparison confirms that \textsc{delphic} outperforms the traditional approach in terms of both space and time. We point out that time and space gains are obtained in the \emph{average case}, as there exist extreme (\emph{worst}) cases where the two semantics produce epistemic states with the same structure. This follows by the fact that the \textsc{delphic} framework is semantically equivalent to the Kripke-based one (Theorem \ref{th:delphic_eq}). As a result, the plan existence problems of both frameworks have the same complexity.

    Partial evidences of the advantages of adopting possibilities were already experimentally witnessed in \cite{conf/icaps/Fabiano2020}. However, the planning framework therein corresponds only to a fragment of the DEL framework. 
    Indeed, as mentioned above, an actual possibility-based formalization of actions is there absent, in favour of a direct, ad-hoc encoding of the transition functions of three prototypical types of actions described in the action language $\mal$ \cite{journals/corr/Baral2015}, namely \emph{ontic}, \emph{sensing} and \emph{announcements} actions.
    As already mentioned, we overcome this limitation by equipping \textsc{delphic} with eventualities, which we relate to DEL event models. 

    In conclusion, we provide a threefold contribution:
    \begin{inparaenum}[\itshape (i)]
    \item we introduce \textsc{delphic} as a general DEL planning framework based on possibilities; 
    \item we formally show that \textsc{delphic} constitutes an alternative but semantically equivalent framework for epistemic planning, compared to the Kripke-based framework;
    \item we experimentally show that the underlying model employed by \textsc{delphic} indeed offers promising advantages in performance, in terms of both time and space.
    \end{inparaenum}
    
    The paper is organised as follows. In Section~\ref{sec:del}, we recall the necessary preliminaries on DEL; in Section~\ref{sec:delphic}, we formally define \textsc{delphic} and we show its equivalence with the Kripke-based framework 
    and in Section~\ref{sec:eval} we show our experimental evaluation.

    \section{Preliminaries}\label{sec:del}
    In this section we provide the required preliminaries on DEL~\cite{book/springer/vanDitmarsch2007} by illustrating its fundamental components: 
    epistemic models in Section~\ref{sec:epistemic_models}, event models in Section~\ref{sec:event_models}, and the product update in Section~\ref{sec:product_update}. 
    Although the notion of possibility is part of the preliminaries~\cite{journals/jolli/Gerbrandy1997}, we defer these to Section~\ref{sec:delphic}, as this allows us to illustrate the components of \textsc{delphic} by following a similar structure. 

\subsection{Epistemic Models}\label{sec:epistemic_models}

    Let us fix a countable set $\atomSet$ of propositional atoms  and a finite set $\agentSet = \{1, \dots, n\} $ of agents.  The language $\Lang{}$ of \emph{multi-agent epistemic logic on $ \atomSet $ and $ \agentSet $ with common knowledge/belief} is defined by the following BNF:
        \begin{equation*}
            \varphi ::= \top \mid \atom{p} \mid \neg \varphi \mid \varphi \wedge \varphi \mid \Box_i \varphi,
        \end{equation*}

        \noindent where $ \atom{p} \in \atomSet $, $ i \in \agentSet $, and $ G \subseteq \agentSet $. Formulae of the form $ \Box_i \varphi $ are read as ``agent $ i $ knows/believes that $ \varphi $''. We define the dual operators $\Diamond_i$ as usual. The semantics of DEL formulae is based on the concept of \emph{possible worlds}. \emph{Epistemic models} are defined as \emph{Kripke models}~\cite{journals/apf/Kripke1963} and they contain both factual information about possible worlds and epistemic information, i.e., which worlds are considered possible by each agent.

        \begin{definition}[Kripke Model]
            A \emph{Kripke model} for $ \Lang{} $ is a triple $ M = (W, R, V) $ where:
            \begin{compactitem}
                \item $ W \neq \varnothing $ is the set of possible worlds.
                \item $ R: \agentSet \rightarrow 2^{W \times W} $ assigns  to each agent $ i $ an accessibility relation $ R(i) $.
                \item $ V: \atomSet \rightarrow 2^W $ assigns to each atom a set of worlds.
            \end{compactitem}
        \end{definition}

        \noindent We abbreviate the relations $ R(i) $ with $ R_i $ and use the infix notation $ w R_i v $ in place of $ (w, v) \in R_i $. An \emph{epistemic state} in DEL is defined as a \emph{multi-pointed Kripke model} (\emph{MPKM}), \ie as a pair $ (M, W_d) $, where $ W_d \subseteq W $ is a non-empty set of designated worlds.

        \begin{example}[Coin in the Box]\label{ex:k_model}
            Agents $ a $ and $ b $ are in a room where a box is placed. Inside the box there is a coin. None of the agent knows whether the coin lies heads ($ \atom{h} $) or tails up ($ \neg \atom{h} $). Both agents know the perspective of the other.
            This is represented by the following MPKM (where the circled bullet represent the designated world).

            {
                \centering
                \tikzstyle{world}       =[circle,   thick,draw=black,       fill=black,minimum size=5pt,inner sep=0pt]
\tikzstyle{poinetdworld}=[circle,   thick,draw=black,double,fill=black,minimum size=5pt,inner sep=0pt]

\begin{tikzpicture}[,->,>=stealth',auto,
    thick,style={font=\sffamily\footnotesize\bfseries}]

    \node[poinetdworld, label={left:$w_1{:}h$}]            (w1)  at (0,    0)   {};

    \node[world,        label={right:$w_2{:}\neg h$}]       (w2)  at (1.5,  0)   {};

    \path
        (w1) edge[<->]                                node[above]       {$a,b$}   (w2)
        (w1) edge[loop, out=50, in=130, looseness=14] node[left]  {$a,b~~$} (w1)
        (w2) edge[loop, out=50, in=130, looseness=14] node[right] {$~a,b$}  (w2)
    ;
\end{tikzpicture}

            \par}
        \end{example}

        \begin{definition}[Truth in Kripke Models]\label{def:k-truth}
            Let $ M = (W, R, V) $ be a Kripke model, $ w \in W $, $ i \in \agentSet $, $ \atom{p} \in \atomSet $ and $\varphi,\psi \in \Lang{C}$ be two formulae. Then,

            {\centering
            $
                \begin{array}{@{}lll}
                    (M, w) \models \atom{p}            & \text{ iff } & w \in V(\atom{p}) \\
                    (M, w) \models \neg \varphi        & \text{ iff } & (M, w) \not\models \varphi \\
                    (M, w) \models \varphi \wedge \psi & \text{ iff } & (M, w) \models \varphi \text{ and } (M, w) \models \psi \\
                    (M, w) \models \Box_i \varphi      & \text{ iff } & \forall v \text{ if } w R_i v \text{ then } (M, v) \models \varphi
                \end{array}
            $\par}
            \noindent Moreover, $ (M, W_d) \models \varphi $ iff $ (M, v) \models \varphi $, for all $ v \in W_d $.
        \end{definition}

        We recall the notion of bisimulation for MPKMs \cite{conf/kr/Bolander2021}. 
                
        \begin{definition}[Bisimulation]
            A \emph{bisimulation} between MPKMs $ ((W, R, V), W_d) $ and $ ((W', R', V'), $ $ W'_d) $ is a binary relation $ B \subseteq W \times W' $ satisfying:
            \begin{compactitem}
                \item \emph{Atoms}: if $ (w, w') \in B $, then for all $ \atom{p} \in \atomSet $, $ w \in V(\atom{p}) $ iff $ w' \in V'(\atom{p}) $.
                \item \emph{Forth}: if $ (w, w') \in B $ and $ w R_i v $, then there exists $ v' \in W' $ such that $ w' R'_i v' $ and $ (v, v') \in B $.
                \item \emph{Back}: if $ (w, w') \in B $ and $ w' R'_i v' $, then there exists $ v \in W $ such that $ w R_i v $ and $ (v, v') \in B $.
                \item \emph{Designated}: if $ w \in W_d $, then there exists a $ w' \in W'_d $ such that $ (w, w') \in B $, and vice versa.
            \end{compactitem}
        \end{definition}

        \noindent We say that two MPKMs $s$ and $s'$ are \emph{bisimilar} (denoted by $s \bisim s'$) when there exists a bisimulation between them. It is well known that bisimilar states satisfy the same formulae, hence encode the same information. 

    \subsection{Event Models}\label{sec:event_models}
        In DEL, actions are modeled by \emph{event models}~\cite{conf/tark/Batlag1998,book/aup/vanDitmarsch2008}, which capture action preconditions and effects from the perspectives of multiple agents at once. Intuitively, \emph{events} represent possible outcomes of the action, accessibility relations describe which events are considered possible by agents, preconditions capture the applicability of events, and postconditions specify how events modify worlds.
        \begin{definition}[Event Model]
            An \emph{event model} for $ \Lang{} $ is a quadruple $ \E = (E, Q, pre, post) $ where:
            \begin{compactitem}
                \item $ E \neq \varnothing $ is a finite set of events.
                \item $ Q: \agentSet \rightarrow 2^{E \times E} $ assigns to each agent $ i $ an accessibility relation $ Q(i) $.
                \item $ pre: E \rightarrow \Lang{} $ assigns to each event a \emph{precondition}.
                \item $ post: E {\rightarrow} (\atomSet {\rightarrow} \Lang{}) $ assigns to each event a \emph{postcondition} for each atom.
            \end{compactitem}
        \end{definition}

        \noindent We abbreviate $ Q(i) $ with $ Q_i $ and use the infix notation $ e Q_i f $ in place of $ (e, f) \in Q_i$. An \emph{epistemic action}\footnote{We use ``epistemic action'' with a broad meaning, simply referring to actions in epistemic planning, irrespective of their effects.} in DEL is defined as a \emph{multi-pointed event model} (\emph{MPEM}), \ie as a pair $ (\E, E_d) $, where $ E_d \subseteq E $ is a non-empty set of designated events. An action is \emph{purely epistemic} if, for each $ e \in E $, $ post(e) $ is the identity function $ id $; otherwise it is \emph{ontic}.

        \begin{example}\label{ex:e_model}
            Suppose that, in the scenario of Example~\ref{ex:k_model}, agent $ a $ peeks inside the box to learn how the coin has been placed while $b$ is distracted. Two events are needed to represent this situation: $ e_1 $ (the designated event) represents the perspective of agent $a$, who is looking inside the box; $ e_2 $ represents the fact that agent $b$ does not know what $a$ is doing.
            In the figure below, a pair $ \langle pre(e), post(e) \rangle $ represents the precondition and the postconditions of event $ e $.
            
            {
                \centering
                \tikzstyle{event}       =[rectangle,thick,draw=black,       fill=black,minimum size=5pt,inner sep=0pt]
\tikzstyle{pointedevent}=[rectangle,thick,draw=black,double,fill=black,minimum size=5pt,inner sep=0pt]

\begin{tikzpicture}[,->,>=stealth',auto,
    thick,style={font=\sffamily\footnotesize\bfseries}]

    \node[pointedevent, label={left:$e_1{:}\langle h{,}id\rangle$}]       (e1)  at (4,    0)   {};
    \node[event,        label={right:$e_2{:}\langle \top{,}id\rangle$}]       (e2)  at (5.5,  0)   {};

    \path
        (e1) edge[->]                                 node[above]       {$b$}    (e2)
        (e1) edge[loop, out=50, in=130, looseness=11] node[left]  {$a~~$}  (e1)
        (e2) edge[loop, out=50, in=130, looseness=11] node[right] {$~a,b$} (e2)
    ;
\end{tikzpicture}

            \par}
        \end{example}

        We give a notion of bisimulation for actions, which will be needed to show an important relationship with our model. 
                
        \begin{definition}[Bisimulation for actions]
            A \emph{bisimulation} between MPEMs $ ((E, Q, pre, post), E_d) $ and $ ((E', Q',$ $pre', post'), E'_d) $ is a binary relation $ B {\subseteq} E \times E' $ satisfying:
            \begin{compactitem}
                \item \emph{Formulae}: if $ (e, e') \in B $, then $pre(e) = pre'(e')$ and, for all $ \atom{p} \in \atomSet $, $post(e)(p) = post'(e')(p)$.
                \item \emph{Forth}: if $ (e, e') \in B $ and $ e Q_i f $, then there exists $ f' \in W' $ such that $ e' Q'_i f' $ and $ (f, f') \in B $.
                \item \emph{Back}: if $ (e, e') \in B $ and $ e' Q'_i f' $, then there exists $ f \in W $ such that $ e Q_i f $ and $ (f, f') \in B $.
                \item \emph{Designated}: if $ e \in E_d $, then there exists a $ e' \in E'_d $ such that $ (e, e') \in B $, and vice versa.
            \end{compactitem}
        \end{definition}

        \noindent We say that two MPEMs $\alpha$ and $\alpha'$ are \emph{bisimilar} (denoted by $\alpha \bisim \alpha'$) when there exists a bisimulation between them.
        
    \subsection{Product Update}\label{sec:product_update}
        The product update of a MPKM with a MPEM results into a new MPKM that contains the updated information of agents. Here we adapt the definition of van Ditmarsch and Kooi \cite{book/aup/vanDitmarsch2008} to deal with multi-pointed models. An MPEM $ (\E, E_d) $ is \emph{applicable} in $ (M, W_d) $ if for each world $ w \in W_d $ there exists an event $ e \in E_d $ such that $ (M, w) \models pre(e) $.
        \begin{definition}[Product Update]\label{def:update_em}
            The \emph{product update} of a MPKM $ (M, W_d) $ with an applicable MPEM $ (\E, E_d) $, with $ M = (W, R, V) $ and $ \E = (E, Q, pre, post) $, is the MPKM $ (M, W_d) \otimes (\E, E_d) = ((W', R', V'), W'_d) $, where:

            {\centering
            $
            \begin{array}{l@{}l}
                W'           & = \{(w, e) \in W \times E \mid (M, w) \models pre(e)\} \\
                R'_i         & = \{((w, e), (v, f)) \in W' \times W' \mid w R_i v \text{ and } e Q_i f\} \\
                V'(\atom{p}) & = \{(w, e) \in W' \mid (M, w) \models post(e)(\atom{p})\} \\
                W'_d         & = \{(w, e) \in W' \mid w \in W_d \text{ and } e \in E_d\}
            \end{array}
            $\par}
        \end{definition}

        \begin{example}\label{ex:product-update}
            The product update of the MPKM of Example~\ref{ex:k_model} with the MPEM of Example~\ref{ex:e_model} is the MPKM below, where $v_3=(w_1,e_1)$, $v_1=(w_1,e_2)$ and $v_2=(w_2,e_2)$. Now, agent $a$ knows that the coin lies heads up, while $b$ did not change its perspective.
            Importantly, notice that $w_1$ (resp., $w_2$) and $v_1$ (resp., $v_2$) encode the same information, but they are \emph{distinct} objects.
            
            {
                \centering
                \tikzstyle{world}       =[circle,   thick,draw=black,       fill=black,minimum size=5pt,inner sep=0pt]
\tikzstyle{poinetdworld}=[circle,   thick,draw=black,double,fill=black,minimum size=5pt,inner sep=0pt]

\begin{tikzpicture}[,->,>=stealth',auto,
    thick,style={font=\sffamily\footnotesize\bfseries}]

    \node[poinetdworld, label={left :$v_3{:}h$}] (w11) at (9.25,-0.2) {};
    \node[world,        label={above:$v_1{:}h~~~$}] (w12) at (8.5,  0.4) {};
    \node[world,        label={above:$~~~~~v_2{:}\neg h$}] (w22) at (10,   0.4) {};

    \path
        (w11) edge[->]                                 node[left]  {$b$}   (w12)
        (w11) edge[->]                                 node[right] {$b$}   (w22)
        (w11) edge[loop, out=-40, in=35, looseness=14] node[right] {$a$}   (w11)
        (w12) edge[<->]                                node[above] {$a,b$} (w22)
        (w12) edge[loop, out=160, in=240, looseness=14] node[left]  {$a,b$} (w12)
        (w22) edge[loop, out=-60, in=15, looseness=14] node[right] {$a,b$} (w22)
    ;
\end{tikzpicture}

            \par}
        \end{example}

    \subsection{Plan Existence Problem}\label{sec:plan_ex}
        We recall the notions of planning task and plan existence problem in DEL \cite{conf/ijcai/Aucher2013}. 
    
        \begin{definition}[DEL-Planning Task]
        \label{def:planning_task}
            A \emph{DEL-planning task} is a triple $ T = (s_0, \actionSet,$ $ \varphi_g) $, where:
            \begin{inparaenum}[\itshape (i)]
            \item $s_0$ is the initial MPKM; 
            \item $\actionSet$ is a finite set of MPEMs;
            \item $ \varphi_g \in \Lang{C} $ is a \emph{goal formula}.
            \end{inparaenum}
        \end{definition}

        \begin{definition}\label{def:solution}
            A \emph{solution} (or \emph{plan}) to a DEL-planning task $(s_0, \actionSet,$ $ \varphi_g)$ is a finite sequence $ \alpha_1, \dots, \alpha_\ell $ of actions of $\actionSet$ such that:
            \begin{compactenum}
                \item $ s_0 \otimes \alpha_1 \otimes \dots \otimes \alpha_\ell \models \varphi_g $, and
                \item For each $ 1 {\leq} k {\leq} \ell $, $ \alpha_k $ is applicable in $ s_0 \otimes \alpha_1 \otimes \dots \otimes \alpha_{k-1} $.
            \end{compactenum}
        \end{definition}

        \begin{definition}[Plan Existence Problem]\label{def:plan_ex_problem}
            Let $n \geq 1$ and $\mathcal{T}$ be a class of DEL-planning tasks. \planex{$\mathcal{T}$}{$n$} is the following decision problem:
            ``Given a DEL-planning task $ T \in \mathcal{T} $, where $|\agentSet|{=}n$, does $ T $ have a solution?''
        \end{definition}

    \section{\textsc{delphic}}\label{sec:delphic}
    We introduce  the \textsc{delphic} framework for epistemic planning. \textsc{delphic} is built around the concept of \emph{possibility} (Definition \ref{def:possibility}), first introduced by Gerbrandy and Groeneveld to represent epistemic states. We develop a novel representation for epistemic actions inspired by possibilities, which we term \emph{eventualities} (Definition \ref{def:eventuality}). Then, we present a novel characterisation of update, called \emph{union update} (Definition \ref{def:update_pem}), based on possibilities and eventualities.

    \subsection{Possibilities}\label{sec:poss}
        Possibilities are tightly related to \emph{non-well-founded sets}, \ie sets that may give rise to infinite \emph{descents} $X_1 \in X_2 \in \dots$ (\eg $\Omega = \{\Omega\}$ is a n.w.f. set). We refer the reader to Aczel \cite{books/csli/Aczel1988} for a detailed account on non-well-founded set theory.

        \begin{definition}[Possibility]\label{def:possibility}
            A \emph{possibility} $ \poss{u} $ for $ \Lang{} $ is a function that assigns to each atom $ \atom{p} \in \atomSet $ a truth value $ \poss{u}(\atom{p}) \in \{0, 1\} $ and to each agent $ i \in \agentSet $ a \emph{set of possibilities} $ \poss{u}(i) $, called \emph{information state}.
        \end{definition}

        \begin{definition}[Possibility Spectrum]\label{def:p-spectrum}
            A \emph{possibility spectrum} is a finite set of possibilities $ \poss{U} = \{\poss{u}_1, \dots \poss{u}_k\} $ that we call \emph{designated possibilities}.
        \end{definition}

        \noindent Possibility spectrums represent epistemic states in \textsc{delphic} and are able to represent the same information as MPKMs. Intuitively, each possibility $\poss{u}$ represent a possible world and the components $ \poss{u}(\atom{p})$ and $ \poss{u}(i)$ correspond to the valuation function and the accessibility relations of the world, respectively. Finally, the possibilities in a possibility spectrum represent the designated worlds. We formalize this intuition in Proposition \ref{prop:p-comparison}.
        

        \begin{definition}[Truth in Possibilities]\label{def:p-truth}
            Let $ \poss{u} $ be a possibility, $ i \in \agentSet $, $ \atom{p} \in \atomSet $ and $\varphi,\psi \in \Lang{C}$ be two formulae. Then,
            
            {\centering
            $
                \begin{array}{@{}lll}
                    \poss{u} \models \atom{p}            & \text{ iff } & \poss{u}(\atom{p}) = 1 \\
                    \poss{u} \models \neg \varphi        & \text{ iff } & \poss{u} \not\models \varphi \\
                    \poss{u} \models \varphi \wedge \psi & \text{ iff } & \poss{u} \models \varphi \text{ and } \poss{u} \models \psi \\
                    \poss{u} \models \Box_i \varphi      & \text{ iff } & \forall \poss{v} \text{ if } \poss{v} \in \poss{u}(i) \text{ then } \poss{v} \models \varphi
                \end{array}
            $\par}
            \noindent Moreover, $ \poss{U} \models \varphi $ iff $ \poss{v} \models \varphi $, for all $ \poss{v} \in \poss{U} $.
        \end{definition}

        \paragraph{Comparing Possibilities and Kripke Models.}
        Gerbrandy and Groeneveld \cite{journals/jolli/Gerbrandy1997} show how possibilities and Kripke models correspond to each other. In what follows, we extend this result by analyzing the relation between possibility spectrums and MPKMs. First, following \cite{journals/jolli/Gerbrandy1997}, we give some definitions.

        \begin{definition}[Decoration of Kripke Model]\label{def:dec_km}
            The \emph{decoration} of a Kripke model $ M = (W, R, V) $ is a function $ \delta $ that assigns to each world $w \in W$ a possibility $ \poss{w} = \delta(w) $, such that:
            \begin{compactitem}
                \item $ \poss{w}(\atom{p}) = 1 \text{ iff } w \in V(\atom{p}) $, for each $ \atom{p} \in \atomSet $;
                \item $ \poss{w}(i)  = \{\delta(w') \mid w R_i w'\} $, for each $i \in \agentSet $.
            \end{compactitem}
        \end{definition}

        \noindent Intuitively, decorations provide a link between Kripke-based representations and their equivalent possibility-based ones: given $w$ in $M$, the decoration of $M$ returns the possibility that encodes $w$ (its valuation and accessibility relation). 

        \begin{definition}[Picture and Solution]\label{def:pic_km}
            If $ \delta $ is the decoration of a Kripke model $ M = (W, R, V) $ and $ W_d \subseteq W $, then $ (M, W_d) $ is the \emph{picture} of the possibility spectrum $ \poss{W} = \{\delta(w) \mid w \in W_d\} $. $ \poss{W} $ is called \emph{solution} of $ (M, W_d) $.
        \end{definition}

\begin{figure}[t]
            \centering
            \tikzstyle{world}       =[circle,   thick,draw=black,       fill=black,minimum size=5pt,inner sep=0pt]
\tikzstyle{poinetdworld}=[circle,   thick,draw=black,double,fill=black,minimum size=5pt,inner sep=0pt]

\begin{tikzpicture}[,->,>=stealth',auto,thick,
                    outer/.style={draw=gray,dashed,thick}]
    
    \node[outer,inner sep=2pt, label={below:$(M,\{w_1\})$},label={above:Picture}] (A) at (0, 0) {
        \begin{tikzpicture}[,->,solid,>=stealth',auto,thick,style={font=\sffamily\footnotesize\bfseries}]
            \node[poinetdworld, label={below:$w_1$}] (w1)  at (4,    0)   {};
            \node[world,        label={below:$w_2$}] (w2)  at (5.5,  0)   {};

            \path
                (w1) edge[<->]                                (w2)
                (w1) edge[loop, out=50, in=130, looseness=11] (w1)
                (w2) edge[loop, out=50, in=130, looseness=11] (w2)
            ;
        \end{tikzpicture}
    };

    \node[outer,inner sep=6.5pt, label={below:$\poss{W} = \{\poss{w}_1\}$},label={above:Solution}] (B) at (4, 0) {
        \begin{tikzpicture}[,->,solid,>=stealth',auto,thick,style={font=\sffamily\footnotesize\bfseries}]
            \node[] (w1)  at (0,    0)   {$\poss{w_1}$};
        \end{tikzpicture}
    };

    \path
        (A) edge[->] node[below] {$\delta$} node[above] {Decoration} (B)
    ;

\end{tikzpicture}
            \caption{Relation between picture, decoration and solution.}
            \label{fig:ex4}
        \end{figure}
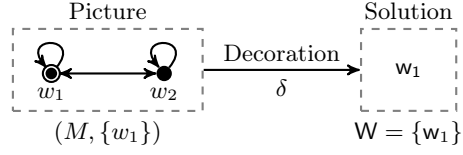

        \noindent Namely, the solution of a MPKM $ (M, W_d) $ is the possibility spectrum $\poss{W}$ that contains the possibilities calculated by the decoration function, one for each designated world in $W_d$. Finally, $ (M, W_d) $ is the picture of $\poss{W}$. Notice that, in general, \emph{different} MPKMs may share the \emph{same} solution. This observation will be formally stated in Proposition \ref{prop:p-comparison}. We now give an example (see also Figure~\ref{fig:ex4}).
        
        \begin{example}\label{ex:dec_km}
            The decoration $ \delta $ of the MPKM of Example~\ref{ex:k_model} assigns the possibilities $ \poss{w}_1 = \delta(w_1) $, $ \poss{w}_2 = \delta(w_2) $. Since $W_d = \{w_1\}$, we have that $ \poss{W} = \{\poss{w}_1\} $ is the solution of $(M,W_d)$, where:
            \begin{compactitem}
                \item $ \poss{w}_1(\atom{h}) = 1 $ and $\poss{w}_1(a) = \poss{w}_1(b) = \{\poss{w}_1, \poss{w}_2\} $;
                \item $ \poss{w}_2(\atom{h}) = 0 $ and $\poss{w}_2(a) = \poss{w}_2(b) = \{\poss{w}_1, \poss{w}_2\} $.
            \end{compactitem}
        \end{example}

        Notice that, in Example~\ref{ex:dec_km}, although the possibility $\poss{w_2}$ is not explicitly part of $\poss{W}$, it is ``stored'' \emph{within} $\poss{w_1}$. That is, we do not lose the information about $\poss{w_2}$.

        Given the above definitions, we are now ready to formally compare possibility spectrums with MPKMs. The following result generalize the one by Gerbrandy and Groeneveld \cite[Proposition 3.4]{journals/jolli/Gerbrandy1997}:

        \begin{proposition}\label{prop:p-comparison}\hfill
            \begin{compactenum}
                \item Each MPKM has a unique decoration;
                \item Each possibility spectrum has a MPKM as its picture;
                \item\label{compact-p} Two MPKMs have the same solution iff they are bisimilar.
            \end{compactenum}
        \end{proposition}

        From item \ref{compact-p} of the above Proposition, we obtain the following remark:

        \begin{remark}\label{rem:compactness}
            Let $s = (M, W_d)$ be a MPKM and let $s'$ be its bisimulation contraction (\ie the smallest MPKM that is bisimilar to $s$). Since $s$ and $s'$ share the same solution $\poss{W}$, it follows that possibility spectrums naturally provide a more compact representation w.r.t. MPKMs.
        \end{remark}
        
		%

        Finally, we show that the solution of a MPKM preserves the truth of formulae.

        \begin{proposition}\label{prop:truth}
            Let $(M, W_d)$ be a MPKM and let $\poss{W}$ be its solution. Then, for every $\varphi \in \Lang{}$, $(M,W_d) \models \varphi$ iff $\poss{W} \models \varphi$.

            \begin{proof}
                Let $\delta$ be the decoration of $(M, W_d)$. We denote with $eq(\psi)$ the fact that $(M,w) \models \psi$ iff $\delta(w) \models \psi$, for all $w {\in} W$.
                
                Consider now $w \in W$ and let $\poss{w} = \delta(w)$. We only need to show that $eq(\varphi)$ holds for any $\varphi \in \Lang{}$. The proof is by induction of the structure of $\varphi$. For the base case, let $\varphi=\atom{p}$. By Definition \ref{def:dec_km}, we immediately have that, for any $\atom{p} \in \atomSet$ and $w \in W$, $(M, w) \models \atom{p}$ iff $\poss{w} \models \atom{p}$ (\ie $eq(\atom{p})$).
                For the inductive step, we have:
                \begin{compactitem}
                    \item Let $\varphi {=} \neg \psi$. From $eq(\psi)$ we get $eq(\neg\psi)$;
                    \item Let $\varphi {=} \psi_1 {\wedge} \psi_2$. From $eq(\psi_1)$, $eq(\psi_2)$ we get $eq(\psi_1 {\wedge} \psi_2)$;
                    \item Let $\varphi {=} \Box_i\psi$ and assume $eq(\psi)$. Then we have: 

                    {\centering
                                 $(M, w) {\models} \Box_i\psi$
                                 $\;\;\textover{\Leftrightarrow}{Def. \ref{def:k-truth}}\;\;$
                                 $\forall v \text{ if } w R_i v, \text{ then } (M, v) {\models} \psi$ \\
                                 $\textover{\Leftrightarrow}{Def. \ref{def:dec_km}, $eq(\psi)$}\quad\;\; $
                                 $\forall \poss{v} \text{ if } \poss{v} \in \poss{u}_i, \text{ then } \poss{v} {\models} \psi$ 
                                 $\;\;\textover{\Leftrightarrow}{Def. \ref{def:p-truth}}\;\;$
                                 $\poss{w} \models \Box_i\psi$
          
                    \par}
                \end{compactitem}
            \end{proof}
        \end{proposition}

    \subsection{Eventualities}
        In \textsc{delphic}, we introduce the novel concept of \emph{eventuality} to model epistemic actions that is compatible with possibilities. In the remainder of the paper, we fix a fresh propositional atom $ \atom{pre} \notin \atomSet $ and let $ \atomSet' = \atomSet \cup \{\atom{pre}\} $. In the following definition, $ \atom{pre} $ encodes the precondition of an event, while the remaining atoms in $ \atomSet $ encode postconditions.

        \begin{definition}[Eventuality]\label{def:eventuality}
            An \emph{eventuality} $ \pem{e} $ for $ \Lang{} $ is a function that assigns to each atom $ \atom{p'} \in \atomSet' $ a formula $ \pem{e}(\atom{p'}) \in \Lang{} $ and to each agent $ i \in \agentSet $ a \emph{set of eventualities} $ \pem{e}(i) $, called \emph{information state}.
        \end{definition}

        \noindent Note that an eventuality is essentially a possibility that associates to each atom a formula (instead of a truth value).

        \begin{definition}[Eventuality Spectrum]\label{def:e-spectrum}
            An \emph{eventuality spectrum} is a finite set of eventualities $ \poss{E} {=} \{\poss{e}_1, \dots \poss{e}_k\} $ that we call \emph{designated eventualities}.
        \end{definition}

        \noindent Eventuality spectrums represent epistemic actions in \textsc{delphic}. 
        Moreover, we can easily show that they are able to represent the same information as MPEMs. 
        Intuitively, each eventuality $\pem{e}$ represents an event and the components $\pem{e}(\atom{pre})$ and $\pem{e}(\atom{p})$ represent the precondition and the postconditions of the event, respectively. Finally, the eventualities in an eventuality spectrum represent the designated events. We formalize this intuition in Proposition \ref{prop:e-comparison}.

    \paragraph{Comparing Eventualities and Event Models.}
        We now analyze the relationship between eventuality spectrums and MPEMs. We introduce the notions of decoration, picture and solution for event models.

        \begin{definition}[Decoration of an Event Model]\label{def:dec_em}
            The \emph{decoration} of an event model $ \E = (E, Q, pre, post) $ is a function $ \delta $ that assigns to each $ e \in E $ an eventuality $ \pem{e} = \delta(e) $, where:
            \begin{compactitem}
                \item $ \pem{e}(\atom{pre}) = pre(e)$ and $ \pem{e}(\atom{p}) = post(e)(\atom{p}) $, for each $\atom{p}\in \atomSet$;
                \item $ \pem{e}(i)          = \{\delta(e') \mid e Q_i e'\} $, for each $i \in \agentSet $.
            \end{compactitem}
        \end{definition}

        \begin{definition}[Picture and Solution]\label{def:pic_em}
            If $ \delta $ is the decoration of an event model $ \E = (E, Q, pre, post) $ and $ E_d \subseteq E $, then $ (\E, E_d) $ is the \emph{picture} of the eventuality spectrum $ \pem{E} = \{\delta(e) \mid e \in E_d\} $ and $ \pem{E} $ is the \emph{solution} of $ (\E, E_d) $.
        \end{definition}

        The above definitions are the counterparts of the notions of decoration and picture given in Definitions~\ref{def:dec_km} and \ref{def:pic_km}. 

        \begin{example}\label{ex:dec_em}
            The decoration $ \delta $ of the MPEM of Example~\ref{ex:e_model} assigns the eventualities $ \pem{e}_1 = \delta(e_1) $ and $ \pem{e}_2 = \delta(e_2) $. Since $E_d = \{e_1\}$, we have that $ \pem{E} = \{\pem{e}_1\} $ is the solution of $(\E,E_d)$, where:
            \begin{compactitem}
                \item $ \pem{e}_1(\atom{pre}) = \atom{h} $; $\pem{e}_1(\atom{h}) = \atom{h}$; $\pem{e}_1(a) {=} \{\pem{e}_1\}$ and $\pem{w}_1(b) {=} \{\pem{w}_2\} $;
                \item $ \pem{e}_2(\atom{pre}) = \top $;     $\pem{e}_2(\atom{h}) = \atom{h}$; $\pem{e}_2(a) = \pem{w}_2(b) = \{\pem{e}_2\} $.
            \end{compactitem}
        \end{example}

        The following results formally compare eventuality spectrums with MPEMs.

        \begin{proposition}\label{prop:e-comparison}\hfill
            \begin{compactitem}
                \item Each MPEM has a unique decoration;
                \item Each eventuality spectrum has a MPEM as its picture;
                \item Two MPEMs have the same solution iff they are bisimilar.
            \end{compactitem}
        \end{proposition}

        Thus, analogously to the case of possibility spectrums, we can see that eventuality spectrums provide us with a compact representation of epistemic actions.
    
    \subsection{Union Update}
        We are now ready to present the novel formulation of update of \textsc{delphic}.  
        We say that an eventuality $\pem{e}$ is \emph{applicable} in a possibility $\poss{u}$ iff $ \poss{u} \models \pem{e}(\atom{pre}) $. Then, an eventuality spectrum $ \pem{E} $ is \emph{applicable} in a possibility spectrums $ \poss{U} $ iff for each $ \poss{u} \in \poss{U} $ there exists an applicable eventuality $ \pem{e} \in \pem{E} $. 

        \begin{definition}[Union Update]\label{def:update_pem}
            The \emph{union update} of a possibility $ \poss{u} $ with an applicable eventuality $ \pem{e} $ is the possibility $ \poss{u'} = \poss{u} \utimes \pem{e} $, where:
            
            {\centering
                $
                \begin{array}{ll}
                    \poss{u'}(\atom{p}) & = 1 \text{ iff } \poss{u} \models \pem{e}(\atom{p}) \\
                    \poss{u'}(i)        & = \{\poss{v} \utimes \pem{f} \mid \poss{v} \in \poss{u}(i), \pem{f} \in \pem{e}(i) \text{ and } \poss{v} \models \pem{f}(\atom{pre})\}
                \end{array}
            $
            \par}
            
            \noindent The \emph{union update} of a possibility spectrum $ \poss{U} $ with an applicable eventuality spectrum $ \pem{E} $ is the possibility spectrum

            {\centering
                $
                \poss{U} \utimes \pem{E} = \{\poss{u} \utimes \pem{e} \mid \poss{u} \in \poss{U}, \pem{e} \in \pem{E} \text{ and } \poss{u} \models \pem{e}(\atom{pre})\}.
                 $
            \par}
        \end{definition}

        \begin{example}\label{ex:union-update}
            The union update of the possibility spectrum $ \poss{W} $ of Example~\ref{ex:dec_km} with the eventuality spectrum of Example~\ref{ex:dec_em} is $\poss{W}\utimes\pem{E}=\{\poss{w_1} \utimes \pem{e_1}\}=\{\poss{v_3}\}$, where $\poss{v_3}(\atom{h})=1$, $\poss{v_3}(a)=\{\poss{v_3}\}$ and $\poss{v_3}(b)=\{\poss{w_1} \utimes \pem{e_2},\poss{w_2} \utimes \pem{e_2}\}=\{\poss{w_1},\poss{w_2}\}$.
            
            Notice that, since $\poss{w_1} \utimes \pem{e_2} {=} \poss{w_1}$ and $\poss{w_2} \utimes \pem{e_2} {=} \poss{w_2}$ the union update allows to reuse previously calculated information.
        \end{example}

    \paragraph{Comparing Union Update and Product Update.}
        Intuitively, it is easy to see that the possibility spectrum of Example \ref{ex:union-update} represents the same information of the MPKM of Example \ref{ex:product-update}. We formalize this intuition with the following lemma, witnessing the equivalence between product and union updates (full proof in the arXiv Appendix).

        \begin{lemma}\label{lem:updates_eq}
            Let $ (\E, E_d) $ be a MPEM applicable in a MPKM $ (M, W_d) $, with solutions $ \pem{E} $ and $ \poss{W} $, respectively. Then the possibility spectrum $ \poss{W'} = \poss{W} \utimes \pem{E} $ is the solution of $ (M', W'_d) = (M, W_d) \otimes (\E, E_d) $.

        \end{lemma}

    \subsection{Plan Existence Problem in \textsc{delphic}}\label{sec:plan_ex_delphic}
        We conclude this section by giving the definitions of planning task and plan existence problem in \textsc{delphic}.
    
        \begin{definition}[\textsc{delphic}-Planning Task]
        \label{def:planning_task_delphic}
            A \emph{\textsc{delphic}-planning task} is a triple $ T = (\poss{W_0}, \Sigma,$ $ \varphi_g) $, where:
            \begin{inparaenum}[\itshape (i)]
            \item $ \poss{W_0} $ is an initial possibility spectrum; 
            \item $ \Sigma $ is a finite set of eventuality spectrums; 
            \item $ \varphi_g \in \Lang{C} $ is a \emph{goal formula}.
            \end{inparaenum}
        \end{definition}

        \begin{definition}\label{def:solution_delphic}
            A \emph{solution} (or \emph{plan}) to a \textsc{delphic}-planning task $(\poss{W_0}, \Sigma,$ $ \varphi_g)$ is a finite sequence $ \pem{E_1}, \dots, \pem{E}_\ell $ of actions of $\Sigma$ such that:
            \begin{compactenum}
                \item $ \poss{W_0} \utimes \pem{E_1} \utimes \dots \utimes \pem{E}_\ell \models \varphi_g $, and
                \item For each $ 1 {\leq} k {\leq} \ell $, $ \pem{E_k} $ is applicable in $ \poss{W_0} \utimes \pem{E_1} \utimes \dots \utimes \pem{E_{k-1}} $.
            \end{compactenum}
        \end{definition}

        \begin{definition}[Plan Existence Problem]\label{def:plan_ex_problem_delphic}
            Let $n {\geq} 1$ and $\mathcal{T}$ be a class of \textsc{delphic}-planning tasks. \planex{$\mathcal{T}$}{$n$} is the following decision problem: ``Given a \textsc{delphic}-planning task $ \poss{T} {\in} \mathcal{T} $, where $|\agentSet|{=}n$, does $ \poss{T} $ have a solution?''
        \end{definition}

        From Lemma \ref{lem:updates_eq}, we immediately get the following result:

        \begin{theorem}\label{th:delphic_eq}
            Let $T = (s_0, \actionSet, \varphi_g)$ be a DEL-planning task and let $\poss{T} = (\poss{W_0}, \Sigma, \varphi_g)$ be a \textsc{delphic}-planning task such that $\poss{W_0}$ is the solution of $s_0$ and $\Sigma$ is the set of solutions of $\actionSet$. Then, $\alpha_1, \dots, \alpha_\ell$ is a plan for $\planex{T}{n}$ iff $\pem{E_1}, \dots, \pem{E}_\ell$ is a plan for $\planex{\poss{T}}{n}$, where $\pem{E_i}$ is the solution of $\alpha_i$, for each $1 \leq i \leq \ell$.
        \end{theorem}

    \section{Experimental Evaluation}\label{sec:eval}
    In this section, we describe our experimental evaluation of the Answer Set Programming (ASP) encodings of \textsc{delphic} and of the traditional Kripke semantics for DEL. Due to space constraints, we provide a brief overview of the encodings\footnote{The full code and documentation of the ASP encodings are available at \url{github.com/a-burigana/delphic\_asp}.} (the full presentation can be found in the arXiv Appendix).

    The aim of the evaluation is to compare the semantics of \textsc{delphic} and the traditional Kripke-based one in terms of both time and space. We do so by testing the encodings on epistemic planning benchmarks collected from the literature\footnote{Due to space limits, the description of the benchmarks is delegated to the arXiv Appendix. All benchmarks are available at \url{github.com/a-burigana/delphic\_asp}.} (\eg \emph{Collaboration and Communication}, \emph{Grapevine} and \emph{Selective Communication}). Time and space performances are respectively evaluated on the total solving time (given in seconds) and the grounding size (\ie the number of ground ASP atoms) provided by the ASP-solver \emph{clingo} output statistics. We now describe the encodings (Section \ref{sec:encodings}) and discuss the obtained results (Section \ref{sec:results}).

    \subsection{ASP Encodings}\label{sec:encodings}
        Since our goal is to achieve a fair comparison the two semantics, we implemented a baseline ASP encoding for both of them. Although optimizations for both encoding are possible, the baseline implementations are sufficient to show our claim. Towards the goal of a fair and transparent comparison, we opted for a declarative language such as ASP (notice that, as our goal is simply to compare the two baselines, the choice of an alternative declarative language would make little difference). In fact, while imperative approaches would render the comparison less clear, as one would need to delve into opaque implementation details, ASP allows to write the code that is transparent and easy to analyze. In fact, the two ASP encodings are very similar, since the representation of \textsc{delphic} objects (possibility/eventuality spectrums) and DEL objects (MPKMs/MPEMs) closely mirror each other. The only difference is in the two update operators (\ie union update and product update). This homogeneity is instrumental to obtain a fair experimental comparison of the two encodings.
        
        We now briefly describe our encodings, assuming that the reader is familiar with the basics concepts of ASP. The two encodings were developed by following the formal definitions of \textsc{delphic} and DEL objects (possibility/eventuality spectrums and MPKMs/MPEMs) and update operators (union and product update) introduced in the previous sections. To increase the efficiency of the solving and grounding phases, the two encodings make use of the \emph{multi-shot} solving approach provided by the ASP-solver \emph{clingo}, which allows for a fine-grained control over grounding and solving of ASP programs. Specifically, this approach allows one to divide an ASP encoding into sub-programs, then handling grounding and solving of these sub-programs separately. In particular, this technique is useful to implement \emph{incremental solving}, which, at each time step, allows to extend the ASP program in order to look for solutions of increasing size.
        Intuitively, every step mimics a Breadth-First Search over the planning state space: at each time step $\asp{t}$, if a solution is not found (\ie there is no plan of length $\asp{t}$ that satisfies the goal), the ASP program is expanded to look for a longer plan.
        For a detailed introduction on multi-shot ASP, we refer the reader to \cite{journals/tplp/Gebser2019,journals/tplp/Kaminski2023}.

        Finally, to visually witness the compactness that possibility spectrums provide w.r.t. MPKMs (see Remark \ref{rem:compactness}), we exploited the Python API offered by \emph{clingo} to implement a graphical representation of the epistemic states visited by the planner. This provides an immediate way of concretely compare the size of output of the two encodings on a given domain instance. Due to space reasons, we report an example of graphical comparison in the arXiv Appendix.

    \subsection{Results}\label{sec:results}
\begin{figure}[t]
            \centering
            \raisebox{0.16cm}{\subfloat[][Space results]{\includegraphics[width=0.49\textwidth]{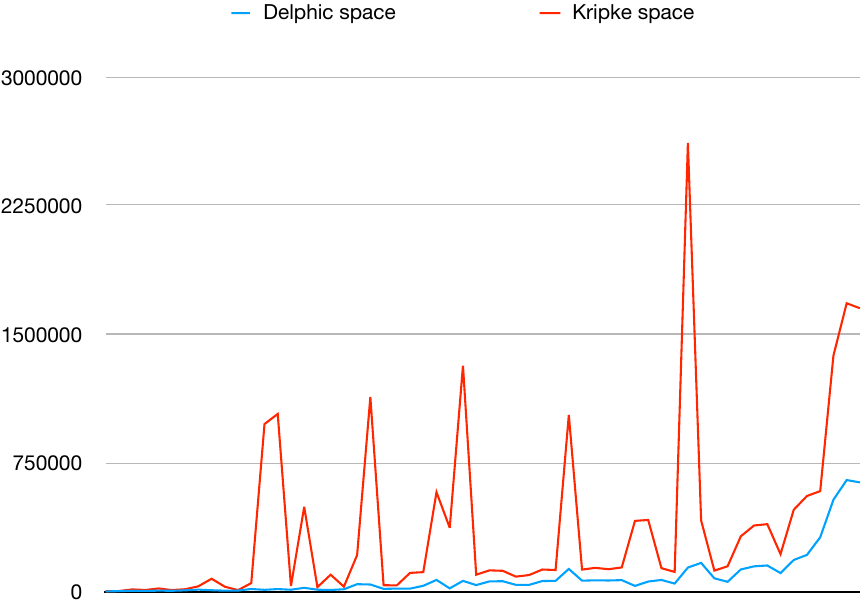}}}
            \hfill
            \subfloat[][Time results]{\includegraphics[width=0.49\textwidth]{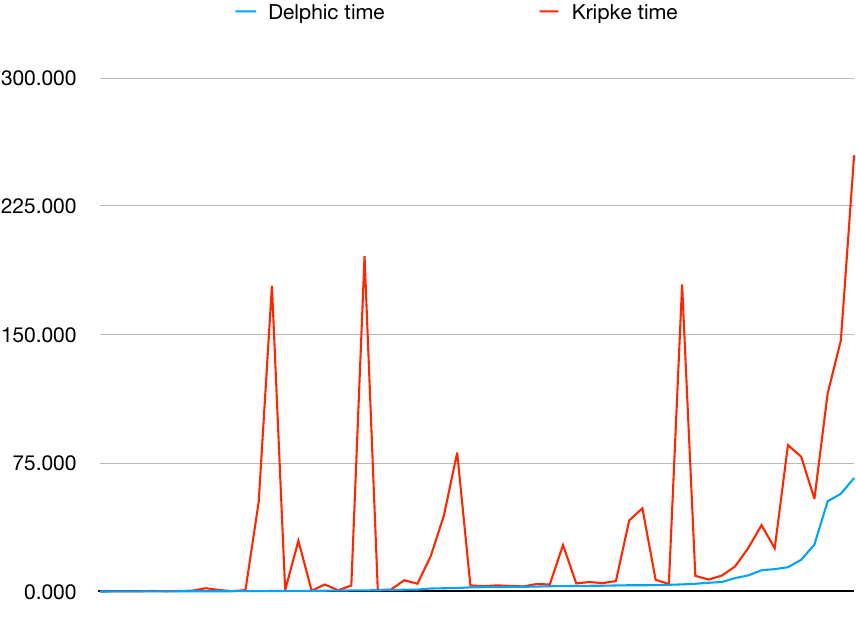}}
            \caption{Results of the evaluation of the \textsc{delphic} and Kripke encodings.}
            \label{fig:results}
        \end{figure}

        We ran our test on a 1.4GHz Quad-Core Intel Core i5 machine with 8GB of memory and with a macOS 12.6 operating system and using \emph{clingo} version 5.6.2 with timeout (t.o.) of 10 minutes. The results are shown in Figure \ref{fig:results}. Space and time results are expressed in number of ASP atoms and in seconds, respectively.
        The comparison clearly shows that the \textsc{delphic} encoding outperforms the one based on the traditional Kripke semantics both in terms of space and time. As shown in Figure \ref{fig:results}.a, the number of ASP atoms produced by the \textsc{delphic} encoding is smaller than the ones produced by the Kripke-based ones. The ``spikes'' witnessed in the latter case are found in presence of instances with longer solutions. This indicates that \textsc{delphic} scales much better in terms of plan length than the traditional Kripke-semantics. In turn, this is positively reflected by the time results graph. In fact, observing space and time results together, we can see how the growth of the size of the epistemic states negatively affects the planning process in terms of time performances. This concretely shows that possibilities can be exploited to achieve more efficient planning tools, thus allowing epistemic planners to be able to deal with the full range of features offered by DEL.
        
\begin{figure}[t]
            \includegraphics[width=\linewidth]{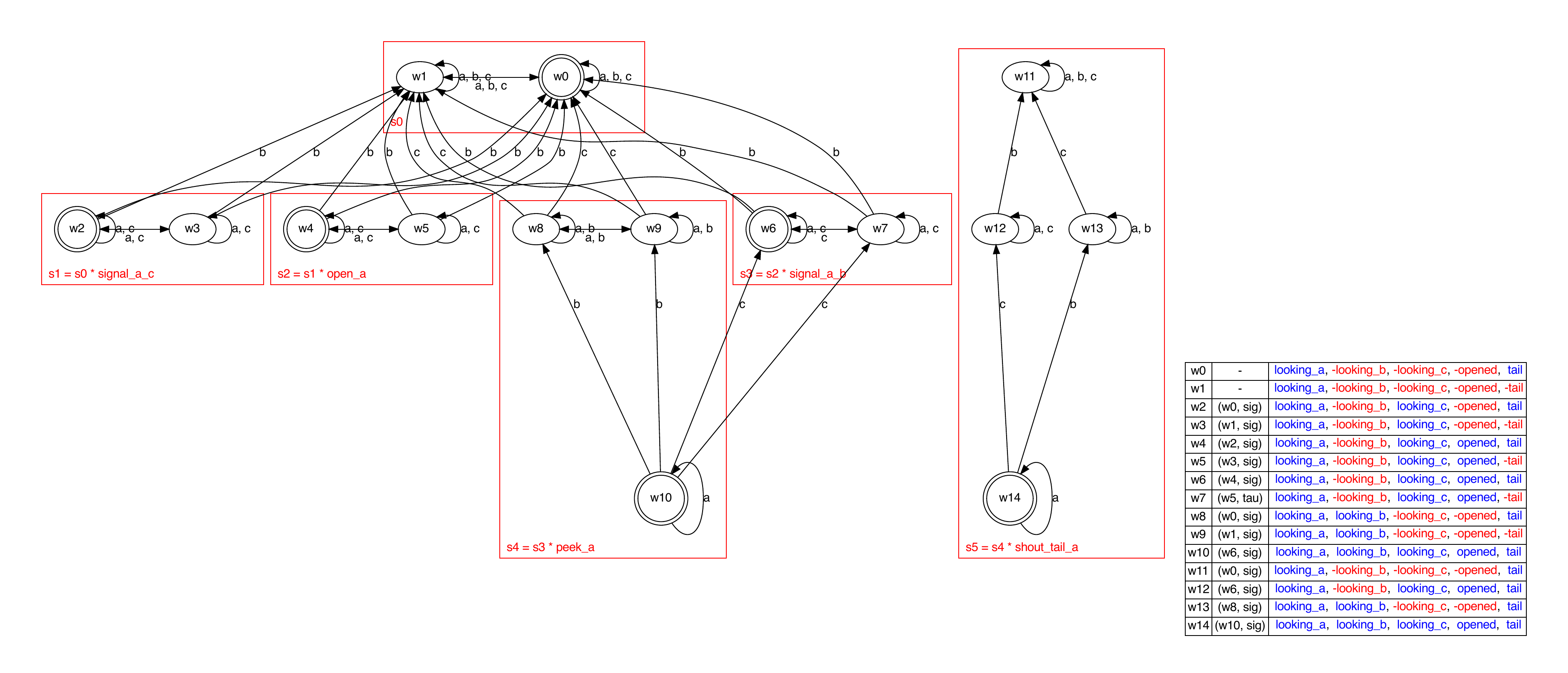}
            \caption{Reuse of previously calculated information in \textsc{delphic}. This figure was obtained by running the terminal command \texttt{python delphic.py -i exp/CB/instance\_\_pl\_5.lp {-}{-}print} (see \url{github.com/a-burigana/delphic\_asp} for the complete documentation).}
            \label{fig:delphic-reuse}
        \end{figure}

        We now analyze the results in detail. The central factor that contributes to the performance gains of \textsc{delphic} is the fact that possibilities allow for a more efficient use of space during the computation of a solution. Specifically, this efficiency results from two key aspects. First, as shown in Remark \ref{rem:compactness}, possibility spectrums are able to represent epistemic information in a more compact way. Working with compact objects contributes significantly to reducing the size of epistemic states after sequences of updates. Second, as shown in Example \ref{ex:union-update}, possibilities naturally allow to reuse previously calculated information (\ie other possibilities that were calculated in previous states). We give a more concrete example of this property in Figure \ref{fig:delphic-reuse}, that shows a sequence of epistemic states (surrounded by rectangles) from a generalization of the Coin in the Box domain of Example \ref{ex:k_model}. We clearly see how the possibilities $\poss{w_0}$ and $\poss{w_1}$ are \emph{reused} in the epistemic states $s_1$, $s_2$, $s_3$ and $s_4$.
        The space efficiency provided by \textsc{delphic} is clearly witnessed in Figure \ref{fig:results}.a. In presence of instances with longer solutions, \textsc{delphic} outperforms the Kripke-based representation, as the latter requires a considerable amount of space to compute a solution (\ie the spikes of the graph).

        The space efficiency of \textsc{delphic} is directly reflected on time performances. Indeed, in Figure \ref{fig:results}.b are shown the same peaks in correspondence of instances with longer solutions. As a result, we can conclude that the \textsc{delphic} framework allows for a more scalable implementation both in terms of space and time performances. Finally, we point out that the analyzed performance gains are obtained in the \emph{average case}, as there exist extreme (\emph{worst}) cases where the two semantics produce epistemic states with the same structure. In fact, we recall that the \textsc{delphic} framework is semantically equivalent to the Kripke-based one (Theorem \ref{th:delphic_eq}). Thus, we can conclude that \textsc{delphic} provides a practical and usable framework for DEL planning that can be exploited to tackle a wide range of concrete epistemic planning scenarios.
        
        We close this section by noting that a similar, but less general result, was obtained by Fabiano et al. \cite{conf/icaps/Fabiano2020}, where a possibility-based semantics is compared to the traditional Kripke-based one on a fragment of DEL called $\mal$ \cite{journals/corr/Baral2015}, that allows three kinds of actions, \ie \emph{ontic}, \emph{sensing} and \emph{announcement} actions. Since \textsc{delphic} is equivalent to the full DEL framework (see Theorem \ref{th:delphic_eq}), our comparison indeed provides a generalization of the claim made by Fabiano et al.
        
    \section{Conclusions}
    We have introduced a novel epistemic planning framework, called \textsc{delphic}, based on the formal notion of possibility, in place of the more traditional Kripke-based DEL representation. We have formally shown that these two frameworks are semantically equivalent. Possibilities provide a more compact representation of epistemic states, in particular by reusing common information across states. To show the benefits of possibilities, we have implemented \textsc{delphic} and the Kripke-based approach in ASP, performing a comparative experimental evaluation with known benchmark domains. The results show that \textsc{delphic} indeed outperforms the Kripke-based approach both in terms of space and time performances, and is thus a good candidate for practical DEL planning.

    In the future, we plan to exploit the performance gains provided by the \textsc{delphic} semantics in more competitive implementations based on C++. An interesting avenue of work is to deepen our analysis of possibility-based succinctness on fragments of DEL, where only a set of specific types of actions are allowed (\eg the language $m\mathcal{A}^*$ \cite{journals/corr/Baral2015} and the framework by Kominis and Geffner \cite{conf/aips/Kominis2015}).
    

    \subsubsection*{Acknowledgements.} This research has been partially supported by the Italian Ministry of University and Research (MUR) under PRIN project PINPOINT Prot. 2020FNEB27, and by the Free University of Bozen-Bolzano with the ADAPTERS project.

    %
    %
    %

    \newpage
    \appendix

    \section{Full Proofs}
    \begin{lemma}\label{lem:updates_eq}
        Let $ (\E, E_d) $ be an MPEM applicable in the MPKM $ (M, W_d) $, with solutions $ \pem{E} $ and $ \poss{W} $, respectively. Then the possibility spectrum $ \poss{W'} = \poss{W} \utimes \pem{E} $ is the solution of $ (M', W'_d) = (M, W_d) \otimes (\E, E_d) $.
        
        \begin{proof}
            Let $ M = (W, R, V) $, $ \E = (E, Q, pre, post) $ and $ M' = (W', R', V') $. Let $ \delta_M $ and $ \delta_{\E} $ be the decorations for $ M $ and $ \E $, respectively. Let then $ (\hat{M}, \hat{W}_d) $ be the picture of $ \poss{W'} $ via the decoration $ \delta $, where $\hat{M} = (\hat{W}, \hat{R}, \hat{V})$. By Proposition 1, to prove that $ \poss{W'} $ is the solution of $ (M', W'_d) $, we need to show that $ (M', W'_d) \bisim (\hat{M}, \hat{W}_d) $. Let $ B \subseteq W' \times \hat{W} $ be a relation such that:
            
            \begin{equation*}
                (w', \hat{w}) \in B \Leftrightarrow w' = (w, e) \wedge \delta(\hat{w}) = \delta_M(w) \utimes \delta_{\E}(e).
            \end{equation*}

            \noindent We now show that $ B $ is a bisimulation between $ M' $ and $ \hat{M} $. Let $ (w', \hat{w}) \in B $, with $ w' = (w, e) $ and let $ v' = (v, f) \in W' $. Let $ \poss{w} = \delta_M(w) $, $ \pem{e} = \delta_{\E}(e) $, $ \poss{v} = \delta_M(v) $ and $ \pem{f} = \delta_{\E}(f) $. Finally, let $ \poss{w'} = \poss{w} \utimes \pem{e} = \delta(\hat{w}) $ and $ \poss{v'} = \poss{v} \utimes \pem{f} $.
            \begin{compactitem}
                \item (Atom) Let $ \atom{p} \in \atomSet $ be a propositional atom. Then:
                
                {\centering
                    \begin{tabular}{lll}
                        $w' \in V'(\atom{p})$ & $\textover{\Leftrightarrow}{Def. 6}$                     & $(M, w) \models post(e)(\atom{p})$ \\
                                              & $\textover{\Leftrightarrow}{Pr.  2, Def. 17}$ & $ \poss{w} \models \pem{e}(\atom{p})$ \\
                                              & $\textover{\Leftrightarrow}{Def. 19}$                    & $\poss{w'}(\atom{p}) = 1 $ \\
                                              & $\textover{\Leftrightarrow}{Def. 14}$                        & $\hat{w} \in \hat{V}(\atom{p})$
                    \end{tabular}
                \par}

                \item (Forth/Back) Let $ i \in \agentSet $ be an agent. Then:
                    
                {\centering
                    \begin{tabular}{lll}
                        $w' R'_i v'$ & $\textover{\Leftrightarrow}{Def. 6}$                     & $w R_i v, e Q_i f, (M, w) \models pre(e) \text{ and }$ \\
                                     &                                                                            & $(M, v) \models pre(f)$ \\
                                     & $\textover{\Leftrightarrow}{Pr.  2, Def. 17}$ & $\poss{v} \in \poss{w}(i), \pem{f} \in \pem{e}(i), \poss{w} \models \pem{e}(\atom{pre}) \text{ and }$ \\
                                     &                                                                            & $\poss{v} \models \pem{f}(\atom{pre})$ \\
                                     & $\textover{\Leftrightarrow}{Def. 19}$                    & $\poss{v'} \in \poss{w}'(i)$ \\
                                     & $\textover{\Leftrightarrow}{Def. 14}$                        & $\hat{w} \hat{R}_i \hat{v}$
                    \end{tabular}
                \par}

                \item (Designated) Let $ (w'_d, \hat{w}_d) \in B $, with $ w'_d = (w, e) $. Then:
                
                {\centering
                    \begin{tabular}{lll}
                        $w'_d {\in} W'_d$ & $\textover{\Leftrightarrow}{Def. 6}$                     & $w {\in} W_d, e {\in} E_d \text{ and } (M, w) {\models} pre(e)$ \\
                                          & $\textover{\Leftrightarrow}{Pr.  2, Def. 17}$ & $\poss{w} \in \poss{W}, \pem{e} \in \pem{E} \text{ and } \poss{w} \models \pem{e}(\atom{pre})$ \\
                                          & $\textover{\Leftrightarrow}{Def. 19}$                    & $\poss{w'} \in \poss{W'}$ \\
                                          & $\textover{\Leftrightarrow}{Def. 14}$                        & $\hat{w}_d \in \hat{W}_d$
                    \end{tabular}
                \par}
            \end{compactitem}
        \end{proof}
    \end{lemma}

    \section{ASP Encodings}\label{sec:asp}
    In this section, we describes the Answer Set Programming (ASP) encodings of \textsc{delphic} and of the traditional Kripke semantics for DEL. We assume that the reader is familiar with the basics concepts of ASP. 
    Notably, our encodings make use of the \emph{multi-shot} solving strategy provided by the ASP-solver \emph{clingo}, which provides fine-grained control over grounding and solving of ASP programs, and is instrumental to implementing an incremental strategy for solving. For a detailed introduction on multi-shot ASP solving, we refer the reader to \cite{journals/tplp/Gebser2019,journals/tplp/Kaminski2023}.
    
    We developed our ASP encodings in such a way that \textsc{delphic} objects (\eg possibility/eventuality spectrums) and DEL objects (\eg MPKMs/MPEMs) are encoded by closely mirroring each other, and only differing in the two update operators (\ie union update and product update). This homogeneity is instrumental to obtain a fair experimental comparison of the two encodings, presented in Section 4 of the main paper. At the same time, as detailed later, we stress that the timings exhibited by our ASP encoding of the Kripke semantics are comparable to those shown by the state-of-the-art solver EFP 2.0 \cite{conf/icaps/Fabiano2020}, which also implements on the Kripke semantics.
    Due to the similarity of the two encodings, in the following we use generic terms such as ``epistemic state/action'' and ``planning task'' to abstract away from which underlying semantics is actually chosen.

    The \textsc{delphic} encoding presented in this section constitutes a contribution of this paper of independent interest. In fact, it generalizes the ASP-based epistemic planner \textsf{PLATO} \cite{journals/tplp/Burigana2020}, which implements a possibility-based semantics for a fragment of DEL \cite{conf/icaps/Fabiano2020}. 
    Moreover, having an ASP-based solver for epistemic planning has several benefits. 
    First, the declarative encoding of the \textsc{delphic} semantics allows for an implementation that is transparent and easier to inspect. 
    Second, we exploited the Python API offered by \emph{clingo} to implement a graphical representation that shows the epistemic states visited by the planner. We thus obtain a practical and useful tool that allows to visualize the evolution of the system. This feature is instrumental in different tasks, such as designing new epistemic planning domains and debugging the correctness of implementations\footnote{The full code and documentation of the ASP encodings are available at 
    \url{https://github.com/a-burigana/delphic\_asp}.}. We concretely show a graphical comparison of the output of the two encodings in the appendix on a concrete domain instance (we do not directly report it here due to space reasons).
    
    The remainder of the section is as follows. First, we briefly describe how \emph{incremental solving} is achieved by the multi-shot solving technique (Section \ref{sec:multi-shot}). Then, we illustrate both encodings component by component: 
    \begin{inparaenum}[\itshape (i)] 
    \item formulae in Section \ref{sec:formuale-enc}, 
    \item planning tasks in Section \ref{sec:task-enc}, 
    \item epistemic states in Section \ref{sec:state-enc}, 
    \item truth conditions in Section \ref{sec:truth-enc}, and 
    \item update operators in Section \ref{sec:update-enc}).
    \end{inparaenum}

    \subsection{Multi-shot Encoding}\label{sec:multi-shot}
        The multi-shot approach allows one to divide an ASP encoding into sub-programs, then handling grounding and solving of these sub-programs separately. In particular, this technique is useful to implement \emph{incremental solving}, which, at each time step, allows to extend the ASP program in order to look for solutions of increasing size. Intuitively, every step mimics a Breadth-First Search over the planning state space: at each time step $\asp{t}$, if a solution is not found (\ie there is no plan of length $\asp{t}$ that satisfies the goal), the ASP program is expanded to look for a longer plan.

        To achieve incremental solving, we build on the approach by Gebser et al. \cite{journals/tplp/Gebser2019}, splitting our encodings in three subprograms:
        \begin{enumerate*}
            \item program $\asp{base}$, which contains all the static information (\ie input information on the planning task);
            \item program $\asp{step(t)}$, where $\asp{t}{>}0$, which describes the evolution of the system (\ie updates semantics); and
            \item program $\asp{check(t)}$, where $\asp{t}{\geq}0$, which verify the truth of formulae of the domain and, in particular, of the goal formula.
        \end{enumerate*}
        Here, $\asp{t}$ represents the current time step that is being considered. In the reminder of this section, when we describe components of the encodings pertaining to the sub-programs $\asp{step(t)}$ and $\asp{check(t)}$, we assume $\asp{t}$ fixed.

    \subsection{Formulae}\label{sec:formuale-enc}
        We represent epistemic formulae through nested ASP predicates. To enhance the performance of grounding and solving, we assume that all input formulae are given in a normal form where occurrences of the $\Box_i$ operators are replaced by the dual representation $\neg\Diamond_i\neg$, and where double negation is simplified.
        Agents and atoms are represented by the ASP predicates $\asp{agent(AG)}$ and $\asp{atom(P)}$, respectively. 
        A formula $\varphi$ is encoded inductively on its structure:
        \begin{enumerate*}
            \item $\varphi=\atom{p}$ is encoded by $\asp{p}$;
            \item $\varphi=\neg\psi$ is encoded by $\asp{neg(PSI)}$;
            \item $\varphi=\psi_1 \wedge \psi_2$ is encoded by $\asp{and(PSI_1, PSI_2)}$; and
            \item $\varphi=\Diamond_i\psi$ is encoded by $\asp{dia(i, PSI)}$.
        \end{enumerate*}

    \subsection{Planning Tasks}\label{sec:task-enc}    
        We now describe our ASP encoding of a planning task.

        \mypar{Initial State.}
        The initial state is given by the following ASP predicates:
        \begin{enumerate*}
            \item $\asp{w\_init(W)}$: $\asp{W}$ is an initial possibility/possible world;
            \item $\asp{r\_init(W_1, W_2, AG)}$: in $\asp{W_1}$, agent $\asp{AG}$ considers $\asp{W_2}$ to be possible;
            \item $\asp{v\_init(W, P)}$: atom $\asp{P}$ is true in $\asp{W}$;
            \item $\asp{dw\_init(W)}$: $\asp{W}$ is a designated possibility/world.
        \end{enumerate*}

        \mypar{Actions.}
        To obtain efficient encodings of the two semantics, and directly support existing benchmarks from the literature, we introduce two features in the definition of actions, namely \emph{(global) action preconditions} and \emph{observability conditions}.

        An action precondition is specified with the ASP predicate $\asp{action\_pre(ACT, PRE)}$ and represents the applicability of the action as a whole. This is syntactic sugar: action preconditions do not modify the expressiveness of epistemic actions, and one can always get an equivalent epistemic action that does not employ them.

        Observability conditions provide a useful way to compactly represent epistemic actions. Namely, agents are split into \emph{observability groups} classifying different perspectives of agents w.r.t.~an action. For instance, in Example 2, 
        agent $a$ is \emph{fully observant}, since it is the one that performs the action, whereas agent $b$ is oblivious, since it does not know that the action is taking place. Then, as we show below, the information regarding which eventuality/event are considered to be possible is lifted to observability groups. Each action must then specify the observability conditions for each agent, assigning each agent to an observability group. This strategy yields an much more succinct representation since each action needs to be substantiated for each group combination, rather than for each agent combination.
        
        We are now ready to describe the ASP encoding of an action $\asp{ACT}$:
        \begin{enumerate*}
            \item $\asp{e(ACT, E)}$: $\asp{E}$ is an eventuality/event of $\asp{ACT}$;
            \item $\asp{q(ACT, E_1, E_2, GR)}$: in $\asp{E_1}$, the observability group $\asp{GR}$ considers $\asp{E_2}$ to be possible;
            \item $\asp{obs(ACT, AG, GR, COND)}$: agent $\asp{AG}$ is in the observability group $\asp{GR}$ if the condition $\asp{COND}$ is satisfied by the current epistemic state;
            \item $\asp{pre(ACT, E, PRE)}$: the precondition of $\asp{E}$ is $\asp{PRE}$;
            \item $\asp{post(ACT, E, P, POST)}$: the postcondition of $\asp{P}$ in $\asp{E}$ is $\asp{POST}$;
            \item $\asp{de(ACT, E)}$: $\asp{E}$ is a designated eventuality/event of $\asp{ACT}$.
        \end{enumerate*}

        We also define some auxiliary ASP predicates that will be used in the update encodings (Section \ref{sec:update-enc}), namely $\asp{idle(ACT, E)}$ and $\asp{inertia(ACT, E, P)}$, which are calculated at the beginning of the ASP computation. The former predicate states that $\asp{E}$ does not affect the worlds in any way (\eg $e_2$ in Example 2 is idle). 
        The latter predicate states that atom $\asp{P}$ is not changed by the postconditions of $\asp{E}$.

        \mypar{Goal.}
        The goal of a planning task is represented by the ASP $\asp{goal(F)}$ predicates. It is possible to declare multiple goal formulae, so that the goal condition of the planning task is the conjunction of all these $\asp{goal}$ predicates.

    \subsection{Epistemic States}\label{sec:state-enc}
        The components of an epistemic state that must be represented (in \textsc{delphic} as well as in the traditional DEL semantics) are four:
        \begin{inparaenum}[\itshape (i)]
            \item possibilities/possible worlds;
            \item information states/accessibility relations;
            \item valuation of propositional atoms;
            \item designated possibilities/worlds.
        \end{inparaenum}

        \mypar{Possibilities and Possible Worlds.}
        To describe possibilities and possible worlds, we make use of the ASP predicate $\asp{w(t, W, E)}$. A possibility/world needs three variable to be univocally identified:
        \begin{compactitem}
            \item $\asp{t}$: represents the time instant when the possibility/world was created. In the initial state, the time is set to $\asp{0}$;
            \item $\asp{W}$: represents the possibility/world that is being updated;
            \item $\asp{E}$: represents the eventuality/event that is updating $\asp{W}$.
        \end{compactitem}

        As at planning time new possibilities/worlds are created dynamically, we are faced with the challenge of finding a suitable ASP representation that correctly and univocally encodes the worlds that are being updated during each action. This is best explained with an example. Suppose we update the possibility/world $\asp{w(t{-}1, W, E)}$ with the eventuality/event $\asp{e(ACT, F)}$ and let $\asp{w(t, X, F)}$ be the result of the update. Intuitively, we can see $\asp{X}$ as representing the world being updated (\ie $\asp{w(t{-}1, W, E)}$). Thus, the challenge we are facing here is to find a suitable representation for $\asp{X}$.

        On the one hand, when encoding possibilities, we have to bear in mind that at each time step, we might end up updating possibilities that were previously calculated at any time in the past. As a result, to correctly and univocally encode $\asp{X}$, we need to keep track of the following information:
        \begin{inparaenum}[\itshape (i)]
            \item the time when a possibility was created;
            \item the identifier of the possibility; and
            \item the eventuality that created it.
        \end{inparaenum}
        As a result, we represent $\asp{X}$ as the ASP tuple $\asp{(t{-}1, W, E)}$.

        On the other hand, when encoding possible worlds, we can always be sure that at each time step we are updating worlds that were created at precisely that time. Thus, the ASP encoding of a possible world is slightly simplified and we represent $\asp{X}$ with the ASP tuple $\asp{(W, E)}$.

        Finally, the initial possibilities/worlds are calculated from the initial state representation with the ASP rule $\asp{w(0, W, null)}$ \texttt{:-} $\asp{w\_init(W).}$, where $\asp{null}$ is a placeholder that indicates that no action occurred before time $\asp{t}$.

        \begin{example}\label{ex:asp-worlds}
            The possibilities of Example 6 are represented in ASP as follows:
            \begin{inparaenum}[\itshape (i)]
                \item $\poss{w_1}$: $\asp{w(0, w_1, null)}$;
                \item $\poss{w_2}$: $\asp{w(0, w_2, null)}$; and
                \item $\poss{v_3}$: $\asp{w(1, (0, w_1, null), e_1)}$.
            \end{inparaenum}
            Again, notice that, in the ASP encoding of possibilities, we are able to reuse previously calculated information.

            Similarly, the possible worlds of Example 3 are represented in ASP as:
            \begin{inparaenum}[\itshape (i)]
                \item $v_1$: $\asp{w(1, (w_1, null), e_2)}$;
                \item $v_2$: $\asp{w(1, (w_2, null), e_2)}$; and
                \item $v_3$: $\asp{w(1, (w_1, null), e_1)}$.
            \end{inparaenum}
        \end{example}

        \mypar{Information States and Accessibility Relations.}
        Let $\asp{w(Tw, W, Ew)}$ and $\asp{w(Tv, V, Ev)}$ be the ASP representations of two possibilities $\poss{w}$ and $\poss{v}$. Since the possibilities contained in the information states of $\poss{w}$ might have been calculated at any time step in the past, in order to encode the fact that $\poss{v} \in \poss{w}(i)$ (where $i \in \agentSet$), we need to keep track of the time when $\poss{v}$ was created. As a result, the resulting encoding is given by the ASP predicate $\asp{r(Tw, W, Ew, Tv, V, Ev, I)}$.

        Let now $\asp{w(Tw, W, Ew)}$ and $\asp{w(Tv, V, Ev)}$ refer to two possible worlds $w$ and $v$. When representing accessibility relations, we can always be sure that when $w R_i v$, the ASP representation of the worlds $w$ and $v$ share the same time value (\ie $\asp{Tw = Tv}$). Thus, we can simplify the encoding as follows: $\asp{r(Tw, W, Ew, V, Ev, I)}$.

        \begin{example}\label{ex:asp-rels}
            The information states of Example 6 
            are represented in ASP as follows (we do not further expand the information states of $\poss{w_1}$ and $\poss{w_2}$ as they refer to previously calculated information):
            \begin{compactitem}
                \item $\poss{v_3} {\in} \poss{v_3}(a)$: $\asp{r(1, (0, w_1, null), e_1, 1, (0, w_1, null), e_1, a)}$;
                \item $\poss{w_1} {\in} \poss{v_3}(b)$: $\asp{r(1, (0, w_1, null), e_1, 0, w_1, null, b)}$; and
                \item $\poss{w_2} {\in} \poss{v_3}(b)$: $\asp{r(1, (0, w_1, null), e_1, 0, w_2, null, b)}$.
            \end{compactitem}

            The accessibility relations of Example 3 
            are represented in ASP as follows:
            \begin{compactitem}
                \item $v_3 R_a v_3$: $\asp{r(1, (w_1, null), e_1, (w_1, null), e_1, a)}$;
                \item $v_3 R_b v_1$: $\asp{r(1, (w_1, null), e_1, (w_1, null), e_2, b)}$;
                \item $v_3 R_b v_2$: $\asp{r(1, (w_1, null), e_1, (w_2, null), e_2, b)}$; and
                \item $v_x R_i v_y$: $\asp{r(1, (w_x, null), e_2, (w_y, null), e_2, i)}$, for each $x,y \in \{1, 2\}$ and $i \in \{a,b\}$.
            \end{compactitem}
        \end{example}

        \mypar{Valuations.}
        For each atom $\asp{P}$, we encode the fact that $\asp{P}$ is true in the possibility/world $\asp{w(t, W, E)}$ with the ASP predicate $\asp{v(t, W, E, P)}$. We only represent \emph{true} atoms. The initial valuation is calculated from the initial state representation with the ASP rule $\asp{v(0, W, null, P)}$ \texttt{:-} $\asp{v\_init(W, P)}$.

        \mypar{Designated Possibilities and Worlds.}
        A designated possibility/world is represented by the ASP predicate $\asp{dw(t, W, E)}$, where variables $\asp{t}$, $\asp{W}$ and $\asp{E}$ have the same meaning as in $\asp{w(t, W, E)}$. The initial designated possibilities/worlds are calculated from the initial state representation with the ASP rule $\asp{dw(0, W, null)}$ \texttt{:-} $\asp{dw\_init(W)}$.

    \subsection{Truth Conditions}\label{sec:truth-enc}
 For space constraints, we abbreviate the representation $\asp{Tx, X, Ex}$ of a possibility/world as $\asp{\bar{X}}$.
        Truth conditions of formulae are encoded by predicate $\asp{holds(t, W, E, F)}$, defined by induction on the structure of formulae as follows:

        {\centering
            \begin{tabular}{l@{}l}
                $\asp{holds(\bar{W}, P)}$             & \texttt{:-~} $\asp{v(\bar{W}, P), atom(P).}$ \\
                $\asp{holds(\bar{W}, neg(F))}$        & \texttt{:-~} $\asp{not\ holds(\bar{W}, F).}$ \\
                $\asp{holds(\bar{W}, and(F_1, F_2))}$ & \texttt{:-~} $\asp{holds(\bar{W}, F_1), holds(\bar{W}, F_2).}$ \\
                $\asp{holds(\bar{W}, dia(AG, F))}$    & \texttt{:-~} $\asp{r(\bar{W}, \bar{V}, AG), holds(\bar{V}, F).}$
            \end{tabular}
        \par}

    \subsection{Update Operators}\label{sec:update-enc}
        We now describe the ASP encodings of the update operators. As the encodings differ, we present them individually. 
        
        \mypar{Union Update.}
        Let the ASP representations of a possibility spectrum $\poss{W}$ at time $\asp{t}$ and of an eventuality spectrum $\pem{ACT}$ be given. We now show how the encoding of the update $\poss{W'} = \poss{W} \utimes \pem{E}$ is obtained. We adopt again the short representation for possibilities/worlds ($\asp{\bar{X}}$). 
        We point out that the following ASP rules have a one-to-one correspondence with Definition 19. 
        First, the designated possibilities of $\poss{W'}$ are determined by the following ASP rules:

        {\centering
            \begin{tabular}{@{}l@{}l@{}l}
                $\asp{dw(t, \bar{W}, E)}$ & \texttt{:-~} & $\asp{dw(\bar{W}), de(ACT, E), pre(ACT, E, PRE),}$ \\
                                          &              & $\asp{holds(\bar{W}, PRE).}$
            \end{tabular}
        \par}

        To describe how possibilities are updated, we need the following ASP predicate:

        {\centering
            \begin{tabular}{@{}l@{}l@{}l}
                $\asp{qt(t, E, F, I)}$   & \texttt{:-~} & $\asp{q(ACT, E, F, GR), obs(t, I, GR).}$
            \end{tabular}
        \par}
        
        \noindent That is, we evaluate the observability conditions of each agent to determine the information states of the action. From this, we can obtain all the updated possibilities recursively:

        {\centering
            \begin{tabular}{@{}l@{}l@{}l}
                $\asp{w(t, \bar{W}, E)}$ & \texttt{:-~} & $\asp{dw(t, \bar{W}, E), de(ACT, E).}$ \\
                $\asp{w(t, \bar{W}, E)}$ & \texttt{:-~} & $\asp{w(t, \bar{V}, F), pre(ACT, E, PRE), \textnormal{-}idle(ACT, E),}$ \\
                                         &              & $\asp{r(\bar{V}, \bar{W}, I), qt(t, F, E, I), holds(\bar{W}, PRE).}$
            \end{tabular}
        \par}

        \noindent The first rule states that a designated possibility is a possibility. Let now $\poss{v'} = \poss{v} \utimes \pem{f}$. Then, the second rule states that for any $\poss{w}$ and $\pem{e}$ such that $\poss{w} \in \poss{v}(i)$, $\pem{e} \in \pem{f}(i)$ and $\poss{w} \models \pem{e}(\atom{pre})$, we create $\poss{w'} = \poss{w} \utimes \pem{e}$. Notice that we require that the eventuality $\pem{e}$ is not \emph{idle}. An eventuality/event is idle when its precondition is $\top$ and when its postconditions are the identity function. In this way, we do not copy redundant information.

        Let now $\asp{\bar{W}'}$ and $\asp{\bar{V}'}$ stand for $\asp{t, \bar{W}, E}$ and $\asp{t, \bar{V}, F}$, respectively, and let $\asp{\bar{U}}$ stand for $\asp{Tu, \bar{U}, Eu}$. We encode information states as follows:

        {\centering
            \begin{tabular}{@{}l@{}l@{}l}
                $\asp{r(\bar{W}', \bar{V}', I)}$ & \texttt{:-~} & $\asp{r(\bar{W}, \bar{V}, I), qt(t, E, F, I).}$ \\
                $\asp{r(\bar{W}', \bar{U}, ~I)}$ & \texttt{:-~} & $\asp{r(\bar{W}, \bar{U}, I), qt(t, E, F, I), Tu{\leq}t, idle(ACT, F).}$
            \end{tabular}
        \par}

        \noindent The first rule states that if $\poss{w'} = \poss{w} \utimes \pem{e}$ and $\poss{v'} = \poss{v} \utimes \pem{f}$ are both created at time $\asp{t}$ and it holds that $\poss{v} \in \poss{w}(i)$ and $\pem{f} \in \pem{e}(i)$, then $\poss{v'} \in \poss{w'}(i)$. The second rule states that if $\poss{w'} = \poss{w} \utimes \pem{e}$ is created at time $\asp{t}$ and it holds that $\poss{u} \in \poss{w}(i)$ and there exists an eventuality $\pem{f}$ such that $\pem{f} \in \pem{e}(i)$ and $\poss{u} = \poss{u} \utimes \pem{f}$, then $\poss{u} \in \poss{w'}(i)$. In this way, we are able to reuse previously calculated information when encoding information states of possibilities.
		
        Finally, 
we encode the valuation of atoms as follows:

        {\centering
            \begin{tabular}{@{}l@{}l@{}l}
                $\asp{v(\bar{W}', P)}$ & \texttt{:-~} & $\asp{post(ACT, E, P, POST), holds(\bar{W}, POST).}$ \\
                $\asp{v(\bar{W}', P)}$ & \texttt{:-~} & $\asp{inertia(ACT, E, P), v(\bar{W}, P).}$
            \end{tabular}
        \par}

        \noindent The first rule states that if $\poss{w} \models \pem{e}(\atom{p})$, then $\poss{w'}(p) {=} 1$. The second rule states that if there are no postconditions associated to an atom (represented by the ASP predicate $\asp{inertia}$), then in $\poss{w'}$ we keep the truth value assigned to $p$ in $\poss{w}$.
        
        \mypar{Product Update.}
            Let the ASP representations of a MPKM $(M, W_d)$ at time $\asp{t}$ and of a MPEM $(\E, E_d)$ be given. The following ASP rules have a one-to-one correspondence with Definition 6. 
            Designated worlds and valuation of atoms are defined in the same way as in the previous case. As above, let $\asp{\bar{W}'}$ and $\asp{\bar{V}'}$ stand for $\asp{t, \bar{W}, E}$ and $\asp{t, \bar{V}, F}$, respectively. Then, updated possible worlds and accessibility relations are encoded as:

            {\centering
                \begin{tabular}{@{}l@{}l@{}l}
                    $\asp{w(t, \bar{W}, E)}$         & \texttt{:-~} & $\asp{w(\bar{W}), e(ACT, E), holds(\bar{W}, PRE).}$ \\
                    $\asp{r(\bar{W}', \bar{V}', I)}$ & \texttt{:-~} & $\asp{r(\bar{W}, \bar{V}, I), qt(t, E, F, I).}$
                \end{tabular}
            \par}

    \subsection{Epistemic Planning Domains}\label{sec:domains}
    We overview the planning domains used for the evaluation.

    \mypar{Assemble Line (AL):} there are two agents, each responsible for processing a different part of a product. Each agent can fail in processing its part and can inform the other agent of the status of her task. The agents can decide to \emph{assemble} the product or to \emph{restart} the process, depending on their knowledge about the product status. This domain is parametrized on the maximum \emph{modal depth} $d$ of formulae that appear in the action descriptions. The goal in this domain is fixed, \ie the agents must assemble the product. The aim of this domain is to analyze the impact of the modal depth both in terms of grounding and solving performances.
    
    \mypar{Coin in the Box (CB):} this is a generalization of the domain presented in Example 1. 
    Three agents are in a room where a closed box contains a coin. None of them knows whether the coin lies heads or tails up. To look at the coin, the agents need to first open the box. Agents can be attentive or distracted: only attentive agents are able to see what actions are executed. Moreover, agents can \emph{signal} others to make them attentive and can also distract them. The goal is for one or more agents to learn the coin position.
    
    \mypar{Collaboration and Communication (CC):} $n {\geq} 2$ agents move along a corridor with $k {\geq} 2$ rooms in which $m {\geq} 1$ boxes are located. Whenever an agent enters a room, it can determine whether a box is there. Agents can then communicate information about the position of the boxes to each other. The goal is to have some agent learn the position of one or more boxes and to know what other agents have learnt.
    
    \mypar{Grapevine (Gr):} $n {\geq} 2$ agents can move along a corridor with $2$ rooms and share their own ``secret'' to agents in the same room. The goal requires agents to know the secret of one or more agents and to hide their secret to others.
    
    \mypar{Selective Communication (SC):} $n {\geq} 2$ agents can move along a corridor with $k {\geq} 2$ rooms. The agents might share some information, represented by an atom $q$. In each room, only a certain subset of agents is able to listen to what others share. The goal requires for some specific subset of agents to know the information and to hide it to others.
    


    \section{Graphical Comparison}
\begin{figure*}
        \includegraphics[width=\linewidth]{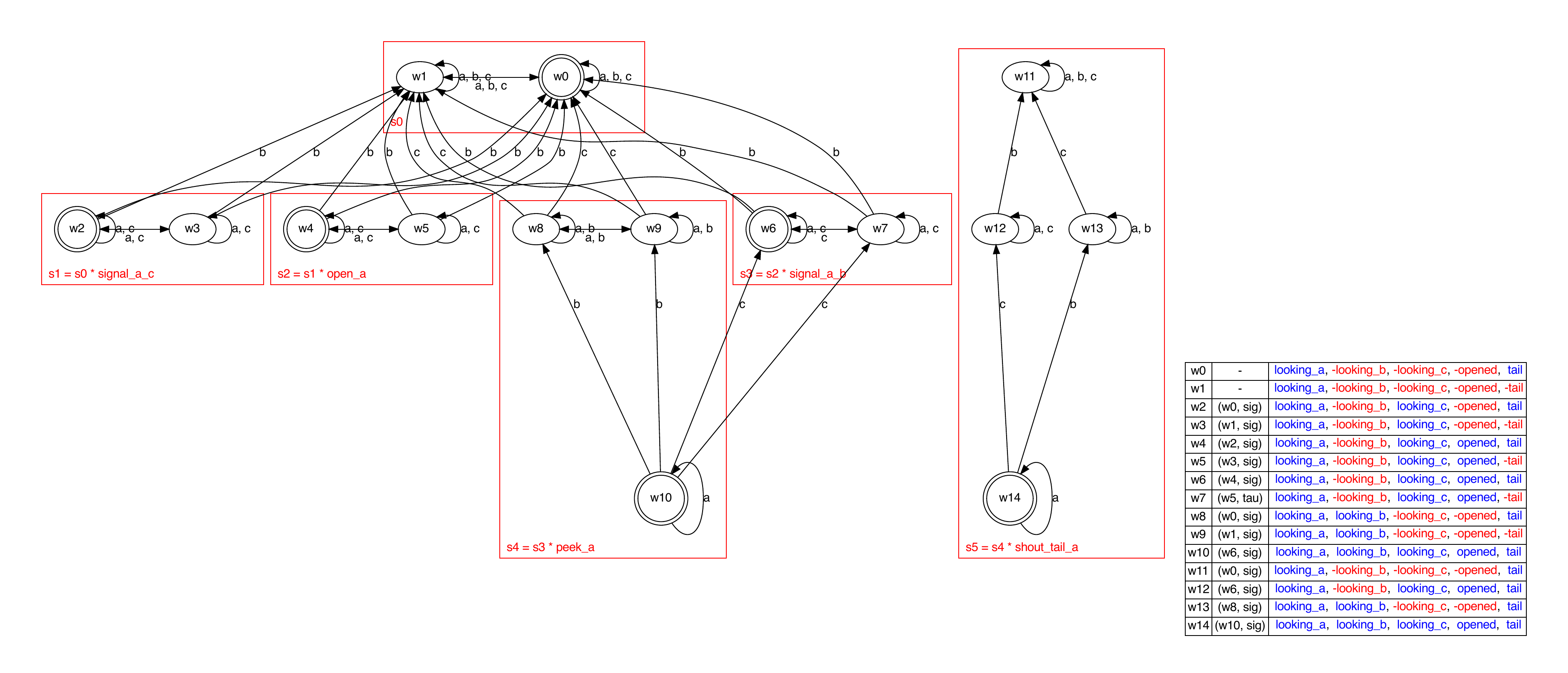}
        \caption{Graphical representation of a sequence of 5 actions using \textsc{delphic}.}
        \label{fig:delphic}
    \end{figure*}

\begin{figure*}
        \includegraphics[width=\linewidth]{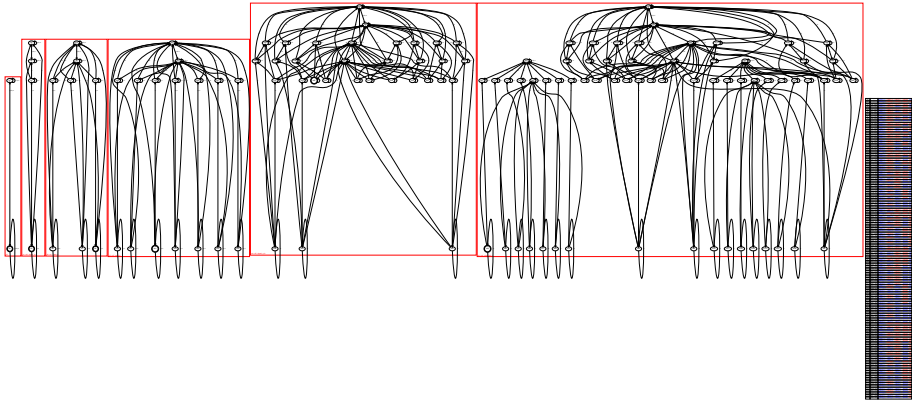}
        \caption{Graphical representation of a sequence of 5 actions using Kripke semantics.}
        \label{fig:kripke}
    \end{figure*}

    We briefly report a concrete example on the succinctness of the possibility-based representation, compared to the Kripke-based one (Figures \ref{fig:delphic} and \ref{fig:kripke}). The figures were generated by our tool. As you can see, even for small plans of length 5, the \textsc{delphic} semantics provide a much more succinct representation. Specifically, in Figure \ref{fig:delphic} we obtain a total of 15 possibilities, while in Figure \ref{fig:kripke} we obtain 126 possible worlds. This clearly confirms the benefits of the \textsc{delphic} semantics.

    \section{Full Experimental Results}
    We here report the complete tables with the results of our experimental evaluations (Tables \ref{tab:al}-\ref{tab:sc}).

    \begin{table*}[t]
    \centering
    \begin{tabular}{|c|c|c|c|c|c|c|c|c|c|}                                                                                                                                                                 \hline
        \multicolumn{10}{|c|}{Assemble}                                                                                                                                                                 \\ \hline
        \multirow{2}{*}{$|\agentSet|$} & \multirow{2}{*}{$|\atomSet|$}  & \multirow{2}{*}{$|W|$}  & \multirow{2}{*}{$|\actionSet|$}  & \multirow{2}{*}{$L$}  & \multirow{2}{*}{$d$} & \multicolumn{2}{|c|}{Delphic} & \multicolumn{2}{|c|}{Kripke} \\ \cline{7-10}
                            &                     &                     &                     &                     &                    & Time          & Atoms         & Time          & Atoms        \\ \hline
        \multirow{10}{*}{2} & \multirow{10}{*}{4} & \multirow{10}{*}{4} & \multirow{10}{*}{6} & \multirow{10}{*}{5} & 2                  & 2.560         & 59332         & 3.153         & 123780       \\
                            &                     &                     &                     &                     & 3                  & 2.621         & 60390         & 3.606         & 121194       \\
                            &                     &                     &                     &                     & 4                  & 2.913         & 61422         & 4.396         & 128644       \\
                            &                     &                     &                     &                     & 5                  & 3.117         & 62478         & 4.148         & 125708       \\
                            &                     &                     &                     &                     & 6                  & 3.304         & 63564         & 4.774         & 128328       \\
                            &                     &                     &                     &                     & 7                  & 3.410         & 64622         & 4.917         & 130464       \\
                            &                     &                     &                     &                     & 8                  & 3.372         & 65680         & 5.556         & 138372       \\
                            &                     &                     &                     &                     & 9                  & 3.566         & 66738         & 6.161         & 140804       \\
                            &                     &                     &                     &                     & 10                 & 3.739         & 67796         & 6.888         & 136576       \\
                            &                     &                     &                     &                     & 24                 & 13.103        & 108000        & 25.264        & 218362       \\ \hline
    \end{tabular}
    \caption{Results for \textbf{AL}.}
    \label{tab:al}
\end{table*}

    \begin{table*}[t]
    \centering
    \begin{tabular}{|c|c|c|c|c|c|c|c|c|c|}                                                                                                                                                              \hline
        \multicolumn{10}{|c|}{Coin in the Box}                                                                                                                                                       \\ \hline
        \multirow{2}{*}{$|\agentSet|$} & \multirow{2}{*}{$|\atomSet|$} & \multirow{2}{*}{$|W|$} & \multirow{2}{*}{$|\actionSet|$}  & \multirow{2}{*}{$L$} & \multirow{2}{*}{$d$} & \multicolumn{2}{|c|}{\textsc{delphic}} & \multicolumn{2}{|c|}{Kripke} \\ \cline{7-10}
                            &                    &                    &                     &                    &                    & Time          & Atoms         & Time          & Atoms        \\ \hline
        \multirow{5}{*}{3}  & \multirow{5}{*}{5} & \multirow{5}{*}{2} & \multirow{5}{*}{21} & 2                  & 1                  & 0.077         & 2459          & 0.098         & 3094         \\
                            &                    &                    &                     & 3                  & 1                  & 0.215         & 5828          & 0.231         & 8394         \\
                            &                    &                    &                     & 5                  & 3                  & 5.137         & 77310         & 7.014         & 122265       \\
                            &                    &                    &                     & 6                  & 3                  & 27.428        & 316840        & 54.091        & 586037       \\
                            &                    &                    &                     & 7                  & 3                  & t.o.          & -             & t.o.          & -            \\ \hline
    \end{tabular}
    \caption{Results for \textbf{CB}.}
    \label{tab:cb}
\end{table*}

    \begin{table*}[t]
    \centering
    \begin{tabular}{|c|c|c|c|c|c|c|c|c|c|}                                                                                                                                                                \hline
        \multicolumn{10}{|c|}{Collaboration and Communication}                                                                                                                                     \\ \hline
        \multirow{2}{*}{$|\agentSet|$} & \multirow{2}{*}{$|\atomSet|$}  & \multirow{2}{*}{$|W|$}  & \multirow{2}{*}{$|\actionSet|$}  & \multirow{2}{*}{$L$} & \multirow{2}{*}{$d$} & \multicolumn{2}{|c|}{Delphic} & \multicolumn{2}{|c|}{Kripke} \\ \cline{7-10}
                            &                     &                     &                     &                    &                    & Time           & Atoms        & Time          & Atoms        \\ \hline
        \multirow{6}{*}{2}  & \multirow{6}{*}{10} & \multirow{6}{*}{4}  & \multirow{6}{*}{20} & 3                  & 2                  & 0.129          & 4579         & 0.186         & 8859         \\
                            &                     &                     &                     & 4                  & 1                  & 0.455          & 10900        & 0.532         & 25916        \\
                            &                     &                     &                     & 5                  & 2                  & 2.467          & 37882        & 3.599         & 98226        \\
                            &                     &                     &                     & 6                  & 2                  & 9.435          & 147183       & 25.365        & 385343       \\
                            &                     &                     &                     & 7                  & 2                  & 66.278         & 636799       & 254.544       & 1652783      \\
                            &                     &                     &                     & 8                  & 2                  & t.o.           & -            & t.o.          & -            \\ \hline
        \multirow{6}{*}{3}  & \multirow{6}{*}{13} & \multirow{6}{*}{4}  & \multirow{6}{*}{30} & 3                  & 2                  & 0.167          & 7112         & 0.328         & 13532        \\
                            &                     &                     &                     & 4                  & 1                  & 0.774          & 15873        & 0.958         & 37125        \\
                            &                     &                     &                     & 5                  & 2                  & 5.580          & 57033        & 9.321         & 147549       \\
                            &                     &                     &                     & 6                  & 2                  & 18.550         & 214086       & 78.667        & 557746       \\
                            &                     &                     &                     & 7                  & 2                  & 143.178        & 934859       & t.o.          & -            \\
                            &                     &                     &                     & 8                  & 2                  & t.o.           & -            & t.o.          & -            \\ \hline
        \multirow{6}{*}{3}  & \multirow{6}{*}{15} & \multirow{6}{*}{8}  & \multirow{6}{*}{42} & 3                  & 2                  & 0.905          & 17288        & 1.153         & 36040        \\
                            &                     &                     &                     & 4                  & 1                  & 3.939          & 46741        & 4.392         & 114453       \\
                            &                     &                     &                     & 5                  & 2                  & 14.207         & 184241       & 85.423        & 477169       \\
                            &                     &                     &                     & 6                  & 2                  & 114.014        & 760234       & 465.422       & 1963514      \\
                            &                     &                     &                     & 7                  & 2                  & t.o.           & -            & t.o.          & -            \\
                            &                     &                     &                     & 8                  & 2                  & t.o.           & -            & t.o.          & -            \\ \hline
        \multirow{5}{*}{2}  & \multirow{5}{*}{14} & \multirow{5}{*}{9}  & \multirow{5}{*}{28} & 3                  & 2                  & 0.541          & 13456        & 0.723         & 28514        \\
                            &                     &                     &                     & 4                  & 1                  & 2.778          & 39077        & 3.020         & 96399        \\
                            &                     &                     &                     & 5                  & 2                  & 12.418         & 152271       & 38.747        & 393337       \\
                            &                     &                     &                     & 6                  & 2                  & 57.073         & 650494       & 146.581       & 1681652      \\
                            &                     &                     &                     & 7                  & 2                  & t.o.           & -            & t.o.          & -            \\ \hline
        \multirow{5}{*}{2}  & \multirow{5}{*}{17} & \multirow{5}{*}{18} & \multirow{5}{*}{40} & 3                  & 2                  & 2.688          & 39310        & 3.324         & 87272        \\
                            &                     &                     &                     & 4                  & 1                  & 7.849          & 128993       & 14.633        & 322395       \\
                            &                     &                     &                     & 5                  & 2                  & 52.642         & 535713       & 115.531       & 1376539      \\
                            &                     &                     &                     & 6                  & 2                  & t.o.           & -            & t.o.          & -            \\
                            &                     &                     &                     & 7                  & 2                  & t.o.           & -            & t.o.          & -            \\ \hline
    \end{tabular}
    \caption{Results for \textbf{CC}.}
    \label{tab:cc}
\end{table*}

    \begin{table*}[t]
    \centering
    \begin{tabular}{|c|c|c|c|c|c|c|c|c|c|}                                                                                                                                                                \hline
        \multicolumn{10}{|c|}{Grapevine}                                                                                                                                                               \\ \hline
        \multirow{2}{*}{$|\agentSet|$} & \multirow{2}{*}{$|\atomSet|$}  & \multirow{2}{*}{$|W|$}  & \multirow{2}{*}{$|\actionSet|$}  & \multirow{2}{*}{$L$} & \multirow{2}{*}{$d$} & \multicolumn{2}{|c|}{Delphic} & \multicolumn{2}{|c|}{Kripke} \\ \cline{7-10}
                            &                     &                     &                     &                    &                    & Time          & Atoms         & Time          & Atoms        \\ \hline
        \multirow{6}{*}{3}  & \multirow{6}{*}{9}  & \multirow{6}{*}{8}  & \multirow{6}{*}{24} & 2                  & 1                  & 0.104         & 3644          & 0.165         & 9256         \\
                            &                     &                     &                     & 3                  & 1                  & 0.197         & 5385          & 0.892         & 28491        \\
                            &                     &                     &                     & 4                  & 1                  & 0.500         & 9294          & 4.142         & 98558        \\
                            &                     &                     &                     & 5                  & 1                  & 1.934         & 18691         & 44.456        & 372271       \\
                            &                     &                     &                     & 6                  & 2                  & 15.943        & 43315         & t.o.          & -            \\
                            &                     &                     &                     & 7                  & 2                  & 52.829        & 107146        & t.o.          & -            \\ \hline
        \multirow{6}{*}{4}  & \multirow{6}{*}{12} & \multirow{6}{*}{16} & \multirow{6}{*}{40} & 2                  & 1                  & 0.360         & 10684         & 0.810         & 32104        \\
                            &                     &                     &                     & 3                  & 1                  & 1.096         & 17182         & 6.613         & 109032       \\
                            &                     &                     &                     & 4                  & 1                  & 3.694         & 33024         & 41.454        & 412426       \\
                            &                     &                     &                     & 5                  & 1                  & 15.049        & 73698         & t.o.          & -            \\
                            &                     &                     &                     & 6                  & 2                  & 87.727        & 184055        & t.o.          & -            \\
                            &                     &                     &                     & 7                  & 2                  & t.o.          & -             & t.o.          & -            \\ \hline
        \multirow{6}{*}{5}  & \multirow{6}{*}{15} & \multirow{6}{*}{32} & \multirow{6}{*}{60} & 2                  & 1                  & 1.153         & 33600         & 4.593         & 113362       \\
                            &                     &                     &                     & 3                  & 1                  & 3.695         & 58527         & 48.582        & 417771       \\
                            &                     &                     &                     & 4                  & 1                  & 9.469         & 123422        & t.o.          & -            \\
                            &                     &                     &                     & 5                  & 1                  & 77.333        & 298973        & t.o.          & -            \\
                            &                     &                     &                     & 6                  & 2                  & t.o.          & -             & t.o.          & -            \\
                            &                     &                     &                     & 7                  & 2                  & t.o.          & -             & t.o.          & -            \\ \hline
    \end{tabular}
    \caption{Results for \textbf{Gr}.}
    \label{tab:gr}
\end{table*}

    \begin{table*}[t]
    \centering
    \begin{tabular}{|c|c|c|c|c|c|c|c|c|c|}                                                                                                                                                               \hline
        \multicolumn{10}{|c|}{Selective Communication}                                                                                                                                                \\ \hline
        \multirow{2}{*}{$|\agentSet|$} & \multirow{2}{*}{$|\atomSet|$}  & \multirow{2}{*}{$|W|$} & \multirow{2}{*}{$|\actionSet|$}  & \multirow{2}{*}{$L$} & \multirow{2}{*}{$d$} & \multicolumn{2}{|c|}{Delphic} & \multicolumn{2}{|c|}{Kripke} \\ \cline{7-10}
                            &                     &                    &                     &                    &                    & Time          & Atoms         & Time          & Atoms        \\ \hline
        \multirow{4}{*}{3}  & \multirow{4}{*}{5}  & \multirow{4}{*}{2} & \multirow{4}{*}{7}  & 3                  & 2                  & 0.026         & 943           & 0.029         & 1848         \\
                            &                     &                    &                     & 5                  & 1                  & 0.111         & 5451          & 0.354         & 18231        \\
                            &                     &                    &                     & 6                  & 3                  & 0.190         & 7775          & 1.948         & 74916        \\
                            &                     &                    &                     & 8                  & 3                  & 2.063         & 62367         & 81.041        & 1318062      \\ \hline
        \multirow{3}{*}{7}  & \multirow{3}{*}{5}  & \multirow{3}{*}{2} & \multirow{3}{*}{7}  & 5                  & 1                  & 0.188         & 11342         & 0.545         & 30615        \\
                            &                     &                    &                     & 7                  & 2                  & 1.778         & 67906         & 20.629        & 579934       \\
                            &                     &                    &                     & 8                  & 2                  & 4.192         & 140081        & 179.118       & 2617071      \\ \hline
        \multirow{4}{*}{8}  & \multirow{4}{*}{11} & \multirow{4}{*}{2} & \multirow{4}{*}{13} & 9                  & 1                  & 0.236         & 10156         & 52.347        & 976904       \\
                            &                     &                    &                     & 10                 & 2                  & 0.339         & 14519         & 178.486       & 1036341      \\
                            &                     &                    &                     & 14                 & 2                  & 17.766        & 494657        & t.o.          & -            \\
                            &                     &                    &                     & 15                 & 2                  & 16.430        & 481155        & t.o.          & -            \\ \hline
        \multirow{4}{*}{9}  & \multirow{4}{*}{11} & \multirow{4}{*}{2} & \multirow{4}{*}{13} & 6                  & 2                  & 4.542         & 167342        & 9.196         & 414842       \\
                            &                     &                    &                     & 8                  & 2                  & 27.503        & 653357        & t.o.          & -            \\
                            &                     &                    &                     & 9                  & 2                  & 94.207        & 1693676       & t.o.          & -            \\
                            &                     &                    &                     & 12                 & 2                  & t.o.          & -             & t.o.          & -            \\ \hline
        \multirow{4}{*}{9}  & \multirow{4}{*}{11} & \multirow{4}{*}{2} & \multirow{4}{*}{13} & 9                  & 2                  & 0.373         & 21893         & 29.649        & 494173       \\
                            &                     &                    &                     & 10                 & 2                  & 0.723         & 41257         & 195.693       & 1135358      \\
                            &                     &                    &                     & 13                 & 2                  & 1.989         & 95485         & t.o.          & -            \\
                            &                     &                    &                     & 17                 & 2                  & 288.827       & 4023556       & t.o.          & -            \\ \hline
        \multirow{8}{*}{9}  & \multirow{8}{*}{12} & \multirow{8}{*}{2} & \multirow{8}{*}{14} & 4                  & 1                  & 0.084         & 4712          & 0.143         & 12602        \\
                            &                     &                    &                     & 5                  & 1                  & 0.234         & 15034         & 0.848         & 49580        \\
                            &                     &                    &                     & 6                  & 1                  & 0.678         & 43058         & 3.560         & 212899       \\
                            &                     &                    &                     & 7                  & 1                  & 3.165         & 132163        & 27.126        & 1030477      \\
                            &                     &                    &                     & 8                  & 1                  & 15.983        & 393797        & t.o.          & -            \\
                            &                     &                    &                     & 9                  & 1                  & 32.373        & 782700        & t.o.          & -            \\
                            &                     &                    &                     & 10                 & 1                  & 129.964       & 2458577       & t.o.          & -            \\
                            &                     &                    &                     & 11                 & 1                  & 289.298       & 4209828       & t.o.          & -            \\ \hline
    \end{tabular}
    \caption{Results for \textbf{SC}.}
    \label{tab:sc}
\end{table*}

\end{document}